\documentclass[journal]{IEEEtran}
\usepackage{graphicx}
\usepackage{multirow}
\usepackage{longtable}
\usepackage{rotating}
\usepackage{color}
\usepackage{makecell}
\usepackage[pagebackref=true,breaklinks=true,letterpaper=true,colorlinks,bookmarks=false]{hyperref}
\usepackage{cite}
\usepackage{lineno}
\usepackage{amsmath}
\usepackage{amssymb}
\graphicspath{{figures/}}

\begin{document}

\title{MDLatLRR: A novel decomposition method for infrared and visible image fusion}

\author{Hui~Li,
        Xiao-Jun~Wu,
	    and Josef Kittler,~\IEEEmembership{Life~Member,~IEEE}
\thanks{This work was supported by the National Key Research and Development Program of China (Grant No. 2017YFC1601800), the National Natural Science Foundation of China (61672265, U1836218), the 111 Project of Ministry of Education of China (B12018), and the UK EPSRC (EP/N007743/1, MURI/EPSRC/DSTL, EP/R018456/1).}
\thanks{Hui Li and Xiao-Jun~Wu(\emph{Corresponding author}) are with the School of Internet of Things Engineering, Jiangnan University, Wuxi 214122, China. (hui\_li\_jnu@163.com, xiaojun\_wu\_jnu@163.com)}
\thanks{Josef Kittler is with the Centre for Vision, Speech and Signal Processing, University of Surrey, Guildford, GU2 7XH, UK. (e-mail: j.kittler@surrey.ac.uk)}
}

\markboth{}%
{Li \MakeLowercase{\textit{et al.}}: MDLatLRR: A novel decomposition method for infrared and visible image fusion}

\maketitle

\begin{abstract}
Image decomposition is crucial for many image processing tasks, as it allows to extract salient features from source images. A good image decomposition method could lead to a better performance, especially in image fusion tasks. We propose a multi-level image decomposition method based on latent low-rank representation(LatLRR), which is called MDLatLRR. This decomposition method is applicable to many image processing fields. In this paper, we focus on the image fusion task. We develop a novel image fusion framework based on MDLatLRR, which is used to decompose source images into detail parts(salient features) and base parts. A nuclear-norm based fusion strategy is used to fuse the detail parts, and the base parts are fused by an averaging strategy. Compared with other state-of-the-art fusion methods, the proposed algorithm exhibits better fusion performance in both subjective and objective evaluation.
\end{abstract}

\begin{IEEEkeywords}
image fusion, latent low-rank representation, multi-level decomposition, infrared image, visible image.
\end{IEEEkeywords}

\IEEEpeerreviewmaketitle

\section{Introduction}
\label{intro}

\IEEEPARstart{I}{n} the field of multi-sensor image fusion, a common task is to combine infrared and visible images. It arises in many applications, such as surveillance \cite{shrinidhi2018ir}, object detection and target recognition \cite{li2019rgb} \cite{lan2018modality}\cite{jiang2019hierarchical}. The key assumption is that different modalities convey complementary information. The main purpose of image fusion is to generate a single image, which combines the complementary sources of information from multiple images of the same scene \cite{li2017pixel}. The key problem is to extract the salient image content from both modalities and then fuse it to produce enhanced output. Many fusion methods have been proposed for this purpose.
The methods most commonly used for image fusion are multi-scale transforms \cite{chen2020infrared}\cite{vishwakarma2018image} and  representation learning based methods \cite{li2018densefuse}\cite{borsoi2019super}. 

In the multi-scale transform domain, there are many fusion tools, such as the discrete wavelet transform(DWT) \cite{ben2005multiscale}, contourlet transform \cite{yang2010image}, shift-invariant shearlet transform \cite{wang2014eggdd}, quaternion wavelet transform \cite{pang2012multifocus} and nonsubsampled shearlet transform \cite{luo2017image}. These decomposition methods project source images to the frequency domain, which increases the computational complexity. Hence, researchers try to find other methods to process source images without a transform, such as representation learning based methods.

In the representation learning domain, the most common methods are based on sparse representation(SR)\cite{wright2008robust} and dictionary learning\cite{chen2019sparse}\cite{chen2019noise}. For instance, Zong et al. \cite{zong2017medical} proposed a novel medical image fusion method based on SR and Histogram of Oriented Gradients(HOG). There are also many algorithms based on combining SR and other tools including pulse coupled neural network(PCNN) \cite{lu2014infrared}, low-rank representation(LRR) \cite{li2017multi} and shearlet transform \cite{yin2017novel}. Other options include the joint sparse representation \cite{liu2017infrared} and cosparse representation \cite{gao2017image}.

Although SR-based fusion methods achieve good fusion performance, these methods are complex, and dictionary learning is a very time-consuming operation, especially online learning. These problems have motivated a growing interest in deep learning to replace the dictionary learning phase in SR.

In deep learning-based fusion methods, deep features of the source images are used to generate the fused image. In \cite{liu2016image}, Liu et al. proposed a fusion method based on convolutional sparse representation(CSR), in which multi-scale and multi-layer features are used to construct the fused output. A follow up work was presented in Liu et al. \cite{liu2017multi}. Image patches which contain different blur versions are used to train a network to produce a decision map. The fused image is obtained by using the decision map and source images. In ICCV 2017, Prabhakar et.al. \cite{prabhakar2017deepfuse} proposed a novel CNN-based fusion framework for a multi-exposure image fusion task.

However, these deep learning-based methods still have drawbacks. The network is difficult to train when the training data is insufficient, especially in infrared and visible image fusion tasks. CNN-based methods tend to work for specific image fusion tasks. To overcome this drawback, Li et.al. \cite{li2018infrared}\cite{li2019infrared} proposed two novel fusion frameworks based on a pretrained network(VGG-19 \cite{simonyan2014very} and ResNet-50 \cite{he2016deep}). The pretrained network is used as  a feature extraction tool. They compared this approach with a strategy involving training a novel network architecture (DenseFuse) \cite{li2018densefuse}. Ma et.al. \cite{ma2019fusiongan} \cite{ma2020infrared}, on the other hand, applied Generative Adversarial Networks to the image fusion task, which also achieves impressive fusion performance. A simple, yet effictive end-to-end fusion framework was proposed by Zhang et. al \cite{zhang2020ifcnn}. Interestingly, in these deep learning based methods, a very little attention is paid to image decomposition.

The latent low rank representation (LatLRR)\cite{liu2011latent} is commonly used in clustering analysis tasks. The authors \cite{liu2011latent} also mentioned that LatLRR can be used to extract salient features from the input data. In this paper we follow this suggestion and propose a novel multi-level decomposition method(MDLatLRR) as a basis of a fusion framework for infrared and visible image fusion.

In our fusion framework, MDLatLRR is used to decompose source images, so as to extract salient detail parts and the base parts. The detail parts are reconstructed by an adaptive fusion strategy and by a reshape operator. The base parts are fused by averaging strategy. Finally, the fused image is constructed by combining the detail parts and the base part. Compared with the state-of-the-art fusion methods, our fusion framework achieves better fusion performance in both subjective and objective evaluations.

The main contributions are summarized as follows,

(1) A LatLRR based multi-level decomposition method (MDLatLRR) is proposed for the image fusion task. In contrast to another LRR-based fusion method~\cite{li2017multi}, which calculates the low-rank coefficients for each image, in our method, the LatLRR problem is only solved once in the training phase to learn a projection matrix $L$. As the size of $L$ is only related to the image patch size,  the fusion method can process the source images of an arbitrary size.

(2) With the multi-level decomposition operation, the proposed fusion method captures more detail information from the source images. As the base part contains preponderance of contour features, it makes the weight-average strategy more effective.

(3) Contrasting it with the $l_1$-norm and $l_2$-norm, the nuclear-norm(the fusion strategy for detail parts) calculates the sum of singular values, which preserves 2D information from the source images. It will generate a larger weight when an input patch contains more texture and structural information. This behaviour is desirable in image processing tasks.

This paper is structured as follows. In Section \ref{pre}, we introduce preliminaries. In Section \ref{mdlatlrr}, we present our multi-level decomposition method(MDLatLRR) in detail. In Section \ref{fusion-method}, the proposed fusion framework based on MDLatLRR is introducted. The experimental results are presented in Section \ref{experiment}. Finally, the conclusions are drawn in Section \ref{conclusions}.


\section{Preliminaries}
\label{pre}

\subsection{Latent Low-Rank Representation}
In 2010, Liu et al. \cite{liu2010robust} proposed the LRR theory, in which the input data matrix is chosen as the dictionary. This method does not achieve good performance when the input data is either insufficient or corrupted. So in 2011, the authors of\cite{liu2011latent} developed the LatLRR theory, according to which the low-rank structure and salient structure are extracted from the raw data. More specifically, the LatLRR problem is formulated as the following optimization problem,
\begin{eqnarray}\label{equ:latlrr}
\begin{split}
  	&\min_{Z,L,E}||Z||_*+||L||_*+\lambda||E||_{1} \\
     &s.t.,X=XZ+LX+E
\end{split}
\end{eqnarray}

\noindent where $\lambda>0$ is a balance coefficient, $||\cdot||_*$ denotes the nuclear norm which is the sum of the singular values of the matrix and $||\cdot||_1$ is $l_1$-norm. $X$ denotes the observed data matrix, $Z$ is a matrix of low-rank coefficients, $L$ is a projection matrix named a salient coefficients matrix, and $E$ is a sparse noisy matrix. Eq \ref{equ:latlrr} is solved by the inexact Augmented Lagrangian Multiplier (ALM) \cite{liu2011latent} algorithm. Then the salient component $LX$ is obtained by Eq \ref{equ:latlrr}. 

\subsection{Projection Matrix $L$ in LatLRR}
\begin{figure}[ht]
\centering
\includegraphics[width=\linewidth]{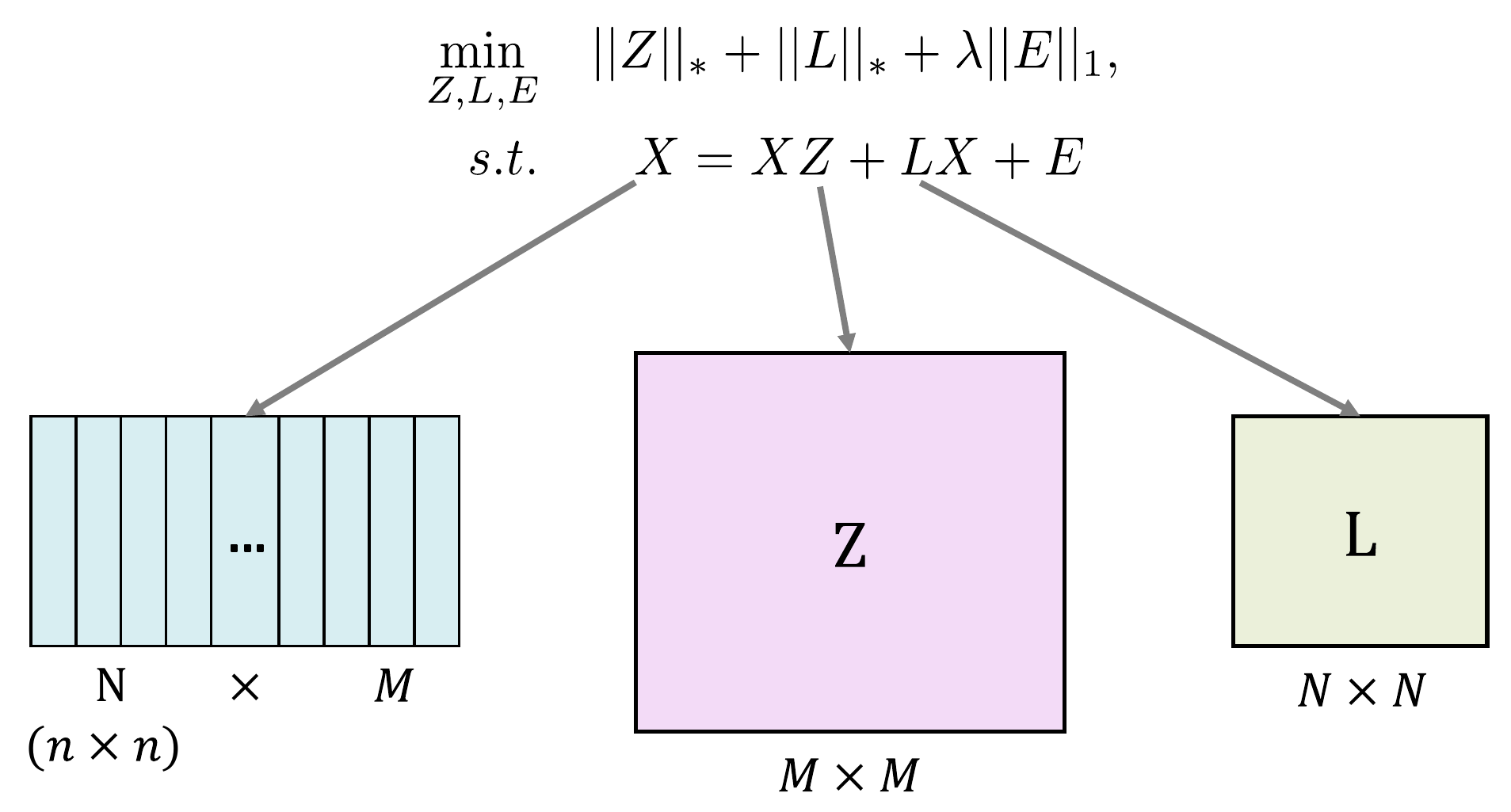}
\caption{Latent low-rank representation. The observed data matrix $X$, low-rank coefficients $Z$ and projection matrix $L$.}
\label{fig:latlrr}
\end{figure}

Assume that a source image is divided into $M$ image patches, as shown in Figure \ref{fig:latlrr}. The size of an image patch is $N=n\times n$. $X$ indicates the observed matrix and each column represents an image patch. The size of $Z$ is related to the number of image patches, which depends on the size of the source image. It is time consuming to calculate the low-rank coefficients for every image in the test phase, if $Z$ is applied to the fusion framework.
In contrast, the size of the projection matrix $L$ is related to the image patch size. In this case, once the projection matrix is learned by LatLRR, it can be used to process other images which are of arbitrary size. In terms of the visual content \cite{liu2011latent}, $LX$ contains more detail information and salient features.


\section{Multi-level Decomposition based on LatLRR(MDLatLRR)}
\label{mdlatlrr}

In this section, our MDLatLRR method is presented. Firstly, we introduce the LatLRR-based decomposition method(DLatLRR). Then, a multi-level version of DLatLRR will be introduced.

\begin{figure}[ht]
\centering
\includegraphics[width=\linewidth]{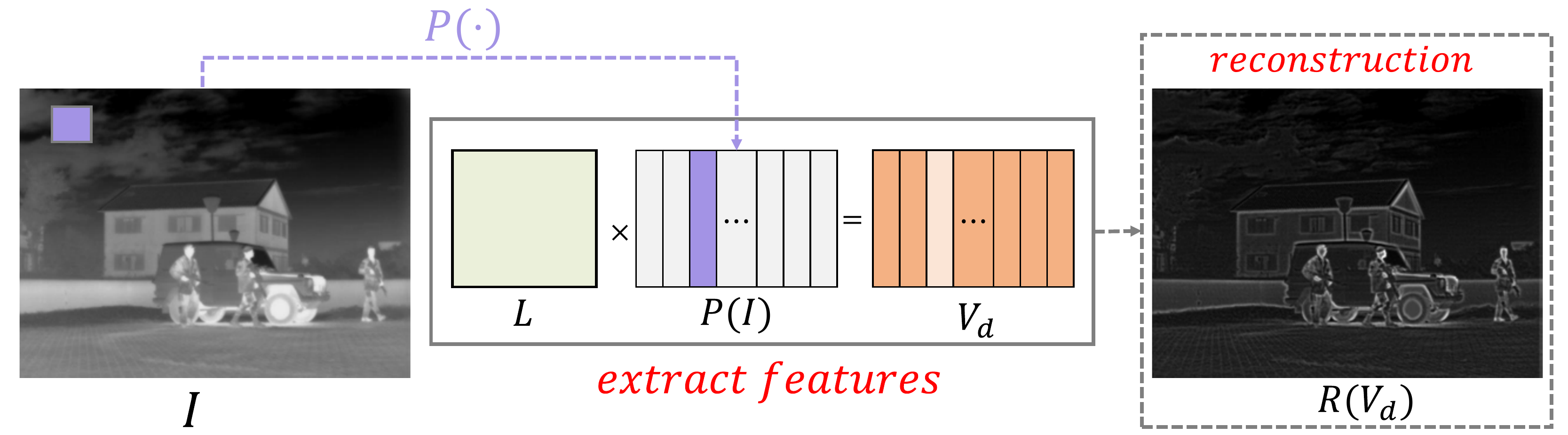}
\caption{The process of image decomposition based on the LatLRR(DLatLRR) method.}
\label{fig:dlatlrr}
\end{figure}

\subsection{DLatLRR}
\label{dlatlrr}

Once the projection matrix $L$ is learned by LatLRR, it can be used to extract the detail part and the base part of the input image by DLatLRR. The process of DLatLRR is illustrated in Figure \ref{fig:dlatlrr}.

\begin{figure*}[ht]
\centering
\includegraphics[width=0.8\linewidth]{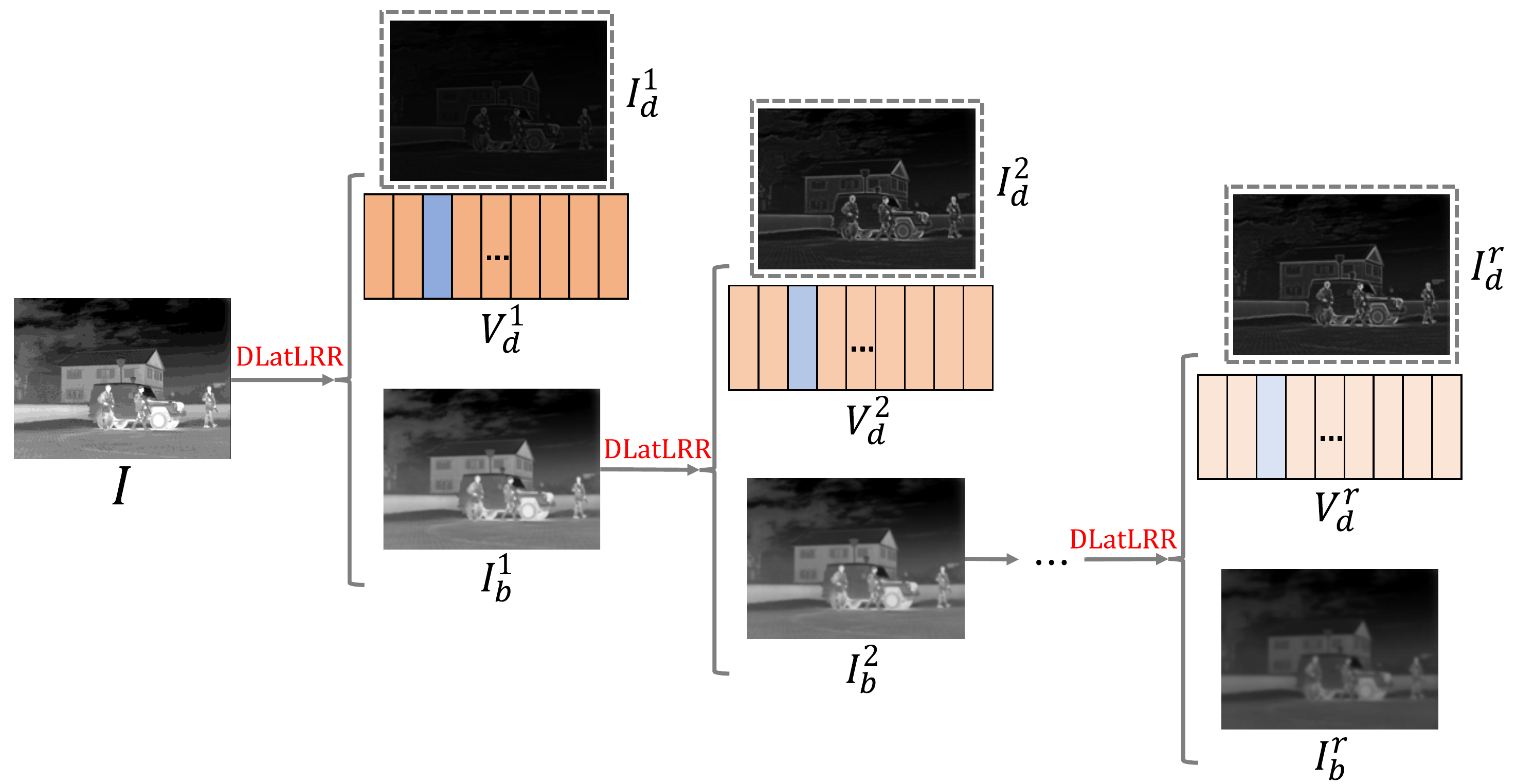}
\caption{MDLatLRR, the deep version of DLatLRR. $r$ denotes the number of decomposition levels.}
\label{fig:mdlatlrr}
\end{figure*}

In Figure \ref{fig:dlatlrr}, the input image($I$) is divided into many image patches by the sliding window technique with an overlap. A sliding window of size $n\times n$ accesses pixel data in the image covered by the window. Stride $s$ means that the window is moved by $s$ pixels in the $x$ and $y$ direction at a time.

These image patches are reshuffled into a source matrix.  Each column of this matrix corresponds to an image patch. The detail part and the base part are determined  by Eq \ref{equ:dlatlrr}.
\begin{eqnarray}\label{equ:dlatlrr}
\begin{split}
  	&V_d=L\times P(I), I_b=I-I_d \\
     &s.t.,I_d=R(V_d)
\end{split}
\end{eqnarray}
\noindent where $L$ denotes the projection matrix learned by LatLRR. $V_d$ is the result of decomposition of the input image($I$), $P(\cdot)$ denotes a two-stage operator which consists of the sliding window technique and reshuffling. $R(\cdot)$ signifies the operator which reconstructs the detail image from the detail part. When restoring $V_d$, the averaging strategy is used to process the overlapping pixel. Firstly, we count the number of the overlapping pixels in each position. Then, for each position in the restored image, we calculate the average pixel value.

As shown in Eq \ref{equ:dlatlrr}, the detail part is generated by $L$, $P(\cdot)$ and the input image $I$. The base part($I_b$) is obtained by subtracting the detail image($I_d$) from the input image.

\subsection{MDLatLRR}

Using the DLatLRR decomposition method, a multi-level decomposition strategy is applied to each base part. As a result, a multi-level version of DLatLRR is obtained, which is named MDLatLRR. The framework of MDLatLRR is presented in Figure \ref{fig:mdlatlrr}.

In Figure \ref{fig:mdlatlrr}, $I$ is the input image, $V_d^i$ and $I_b^i$ denote the detail part matrix and the base part of the input image decomposed by DLatLRR at level $i$. $r$ denotes the highest decomposition level and $i=[1,2,\cdots,r]$. 

As shown in Figure \ref{fig:mdlatlrr}, each previous base part is decomposed by DLatLRR. Then, for $r$ level decomposition, we will get \textbf{r detail part matrices} $V_d^{1:r}$ and \textbf{one base part} $I_b^r$. $I$ can be reconstructed by adding $I_b^r$ and $r$ detail images. The detail parts $V_d^{1:r}$ and base part $I_b^r$ are obtained by Eq \ref{equ:mdlatlrr},
\begin{eqnarray}\label{equ:mdlatlrr}
\begin{split}
  	&V_d^i=L\times P(I_b^{i-1}),\quad I_b^i=I_b^{i-1}-I_d^i \\
     &I_d^i=R(V_d^i),\quad I_b^0=I, i=[1,2,\cdots,r]. 
\end{split}
\end{eqnarray}


\section{MDLatLRR-based Fusion Method}
\label{fusion-method}

In this section, the MDLatLRR-based fusion method is presented in detail. 
\subsection{Learning the Projection Matrix $L$}
\label{learn-l}

\begin{figure}[ht]
\centering
\includegraphics[width=\linewidth]{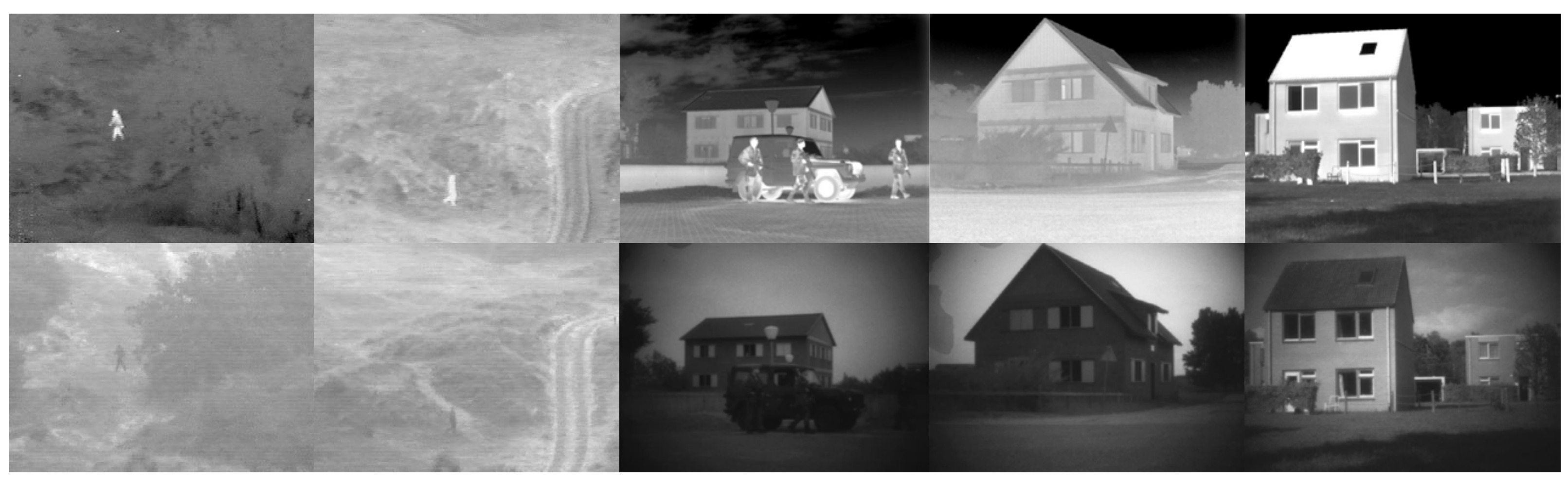}
\caption{Five pairs of infrared and visible images, which are used to learn $L$ by LatLRR.}
\label{fig:training}
\end{figure}

The training images are available at \cite{li2019codes}\footnote{The path is ``/learning\_projection\_matrix/training\_data/''. For more experimental results on extensive training datatset, please refer to our supplementary materials.}. These images are shown in Figure \ref{fig:training}, where the first row contains infrared images and  the second row presents visible images. All the images are divided into image patches by the sliding window technique. 

\begin{figure*}[ht]
\centering
\includegraphics[width=0.95\linewidth]{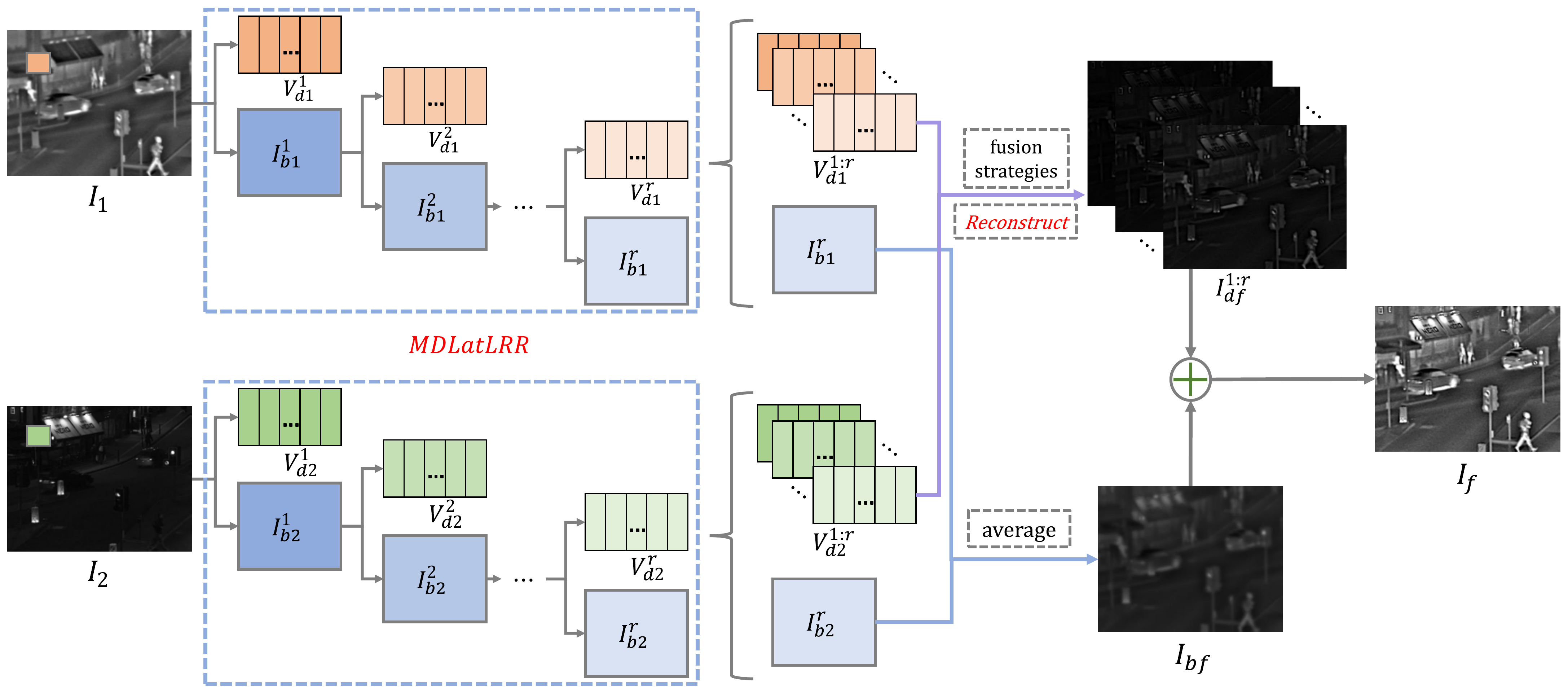}
\caption{The framework of the proposed fusion method. Source images are decomposed into detail parts($V_{dk}^{1:r}$) and base part($I_{bk}^r$). Then, with adaptive fusion strategies, the fused image($I_f$) is reconstructed from fused detail parts($I_{bf}^{1:r}$) and base part($I_{bf}$).}
\label{fig:framework}
\end{figure*}

We obtain 10090 to 40654 image patches, depending on the image patch size, which is $n\times n$ and $n\in{\{8,16\}}$. 2000 image patches are randomly chosen to generate an input matrix $X$ in which each column indicates all pixels of one image patch. The size of $X$ is $N\times M$, where $N=n^2$ and $M=2000$.

As we discussed in Section \ref{pre}, the projection matrix $L$ is learned by LatLRR and ALM. In LatLRR, $\lambda$ is set to 0.4 \cite{liu2011latent}. With different image patch sizes, two types of $L$ ($64\times64,256\times256$) are obtained.

In the fusion framework, $L$ is used to extract salient features from the test images with the same patch size($8\times8,16\times16$). The details of how to train the projection matrix $L$ will be introduced in our ablation study (Section \ref{ablation}).


\subsection{Fusion Method}

Given two input images  $I_1$ and $I_2$ (infrared and visible images), we use $I_k(k\in{\{1,2,\cdots,K\}}, K=2)$ if it is unimportant what type the input image is. The fusion algorithm is the same when the  number of input images $K>2$. The framework of our fusion algorithm is presented in Figure \ref{fig:framework}.

The input image($I_k$) is first decomposed by MDLatLRR to $r$ decomposition levels. Then, $r$ pairs of the detail part matrices($V_{dk}^{1:r}$) and one pair of base part($I_{bk}^r$) are determined. 

For each pair of detail parts, an adaptive fusion strategy is used to fuse these parts column by column. Consequently, $r$ fused detail images($I_{df}^{1:r}$) are obtained. The fused detail images are calculated by Eq \ref{equ:ddlatlrr},
\begin{eqnarray}\label{equ:ddlatlrr}
  	I_{df}^i=R(FS(V_{d1}^i,V_{d2}^i )),i=1,2,\cdots,r
\end{eqnarray}
\noindent where $R(\cdot)$ denotes the reconstruction operator as discussed in Section \ref{dlatlrr}. $FS(\cdot)$ is the fusion strategy which will be introduced in the next subsection.

As the base part contains contour and brightness information, in our fusion method, a weighted average strategy is utilized to obtain the fused base part. 

Once the fused detail images and base part are obtained, the fused image is reconstructed.

In the next subsections, the details of the deep decomposition method, the fusion strategies and the reconstruction process will be presented.

\subsection{Fusion Strategies and Reconstruction}
Once the input image is decomposed by MDLatLRR, we adopt adaptive strategies to fuse these parts.

\subsubsection{Fusion of base parts}

The base parts of input images contain more common features, redundant and brightness information. So, in our fusion method, we use weighted averaging to obtain the fused base part. It is calculated by Eq \ref{equ:fbase},
\begin{eqnarray}\label{equ:fbase}
  	I_{bf}(x,y)=w_{b1} I_{b1}^r(x,y)+w_{b2} I_{b2}^r(x,y)
\end{eqnarray}
\noindent where $(x,y)$ denotes the corresponding position in the base parts($I_{b1}^r$,$I_{b2}^r$) and the fused base part($I_{bf}$). $w_{b1}$ and $w_{b2}$ denote the weights of the base parts.

\subsubsection{Fusion of detail parts}

In contrast to the base parts, the detail parts preserve more structural information, as well as  salient features. So, the fusion strategy($FS(\cdot)$) for detail parts requires a more careful design. 

\begin{figure}[ht]
\centering
\includegraphics[width=\linewidth]{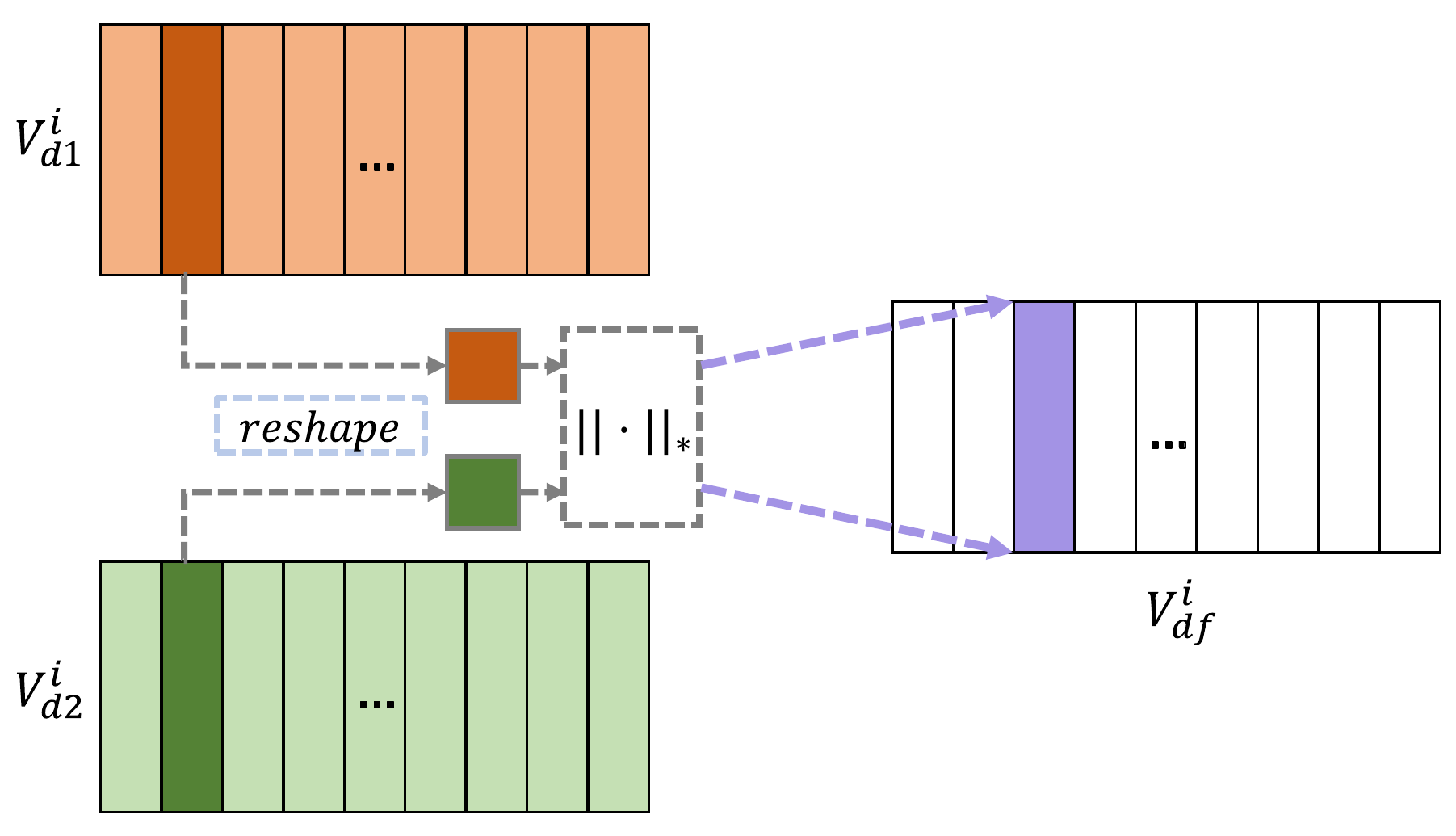}
\caption{Fusion strategy based on nuclear norm. $V_{d1}^i$ and $V_{d2}^i$ denote detail part matrices at the $i$-th level. ``reshape'' means the reconstruction operator($re(\cdot)$) which is used to reconstruct an image patch from the corresponding vectors. $||\cdot||_*$ indicates nuclear-norm. $V_{df}^i$ denotes the fused detail part.}
\label{fig:nunorm}
\end{figure}

In our method, the nuclear norm is used to calculate the weights for each pair of corresponding image patches which are obtained by $P(\cdot)$ from the input images. An overview of this fusion strategy is shown in Figure \ref{fig:nunorm}.

In Figure \ref{fig:nunorm}, $V_{dk}^{i,j}$ and $V_{df}^{i,j}$ are the vectors corresponding to the $j$-th column in the detail part matrix $V_{dk}^i$ and the fused detail part matrix $V_{df}^i$, respectively. $i$ is the decomposition level, $k\in{\{1,2\}}$. 

Firstly, for each corresponding column, the weights $w_{dk}^{i,j}$ are calculated by Eq \ref{equ:nunorm},
\begin{eqnarray}\label{equ:nunorm}
\begin{split}
  	&w_{dk}^{i,j}=\frac{\hat{w_{dk}^{i,j}}}{\sum_{q=1}^K\hat{w_{dq}^{i,j}}} \\
	&\hat{w_{dk}^{i,j}}=||re(V_{dk}^{i,j})||_*
\end{split}
\end{eqnarray}
\noindent where $re(\cdot)$ denotes the operator which is used to reconstruct the image patch from vector $V_{dk}^{i,j}$, and $||\cdot||_*$ indicates the nuclear-norm which calculates the sum of singular values of the matrix. $K$ is the number of input images.

Then, the fused detail part vector $V_{df}^{i,j}$ is obtained by weighting the detail part vectors $V_{dk}^{i,j}$, as shown in Eq \ref{equ:fvector},
\begin{eqnarray}\label{equ:fvector}
  	V_{df}^{i,j}=\sum_{k=1}^K w_{dk}^{i,j}\times V_{dk}^{i,j}
\end{eqnarray}

This fusion strategy is applied to all pairs of detail part matrices $V_{dk}^{1:r}$. Each fused detail image($I_{df}^i$) is calculated by Eq \ref{equ:fdetailim},
\begin{eqnarray}\label{equ:fdetailim}
  	I_{df}^i=R(V_{df}^i),\quad i=[1,2,\cdots,r]
\end{eqnarray}
\noindent where $R(\cdot)$ indicates the reconstruction operator.

\subsubsection{Reconstruction}
\label{recon}

Once we have the fused detail images and the fused base part, the fused image is generated by Eq \ref{equ:recons},
\begin{eqnarray}\label{equ:recons}
  	I_f(x,y)=I_{bf}(x,y)+\sum_{i=1}^rI_{df}^i(x,y)
\end{eqnarray}


\section{Experimental Results and Analysis}
\label{experiment}

The aim of our experiments is to provide a supporting evidence for the key ideas underpinning the proposed method, namely the use of the nuclear-norm and multi-level decomposition. We evaluate the fusion performance of our method and compare it to other existing fusion methods.

\begin{figure}[ht]
\centering
\includegraphics[width=\linewidth]{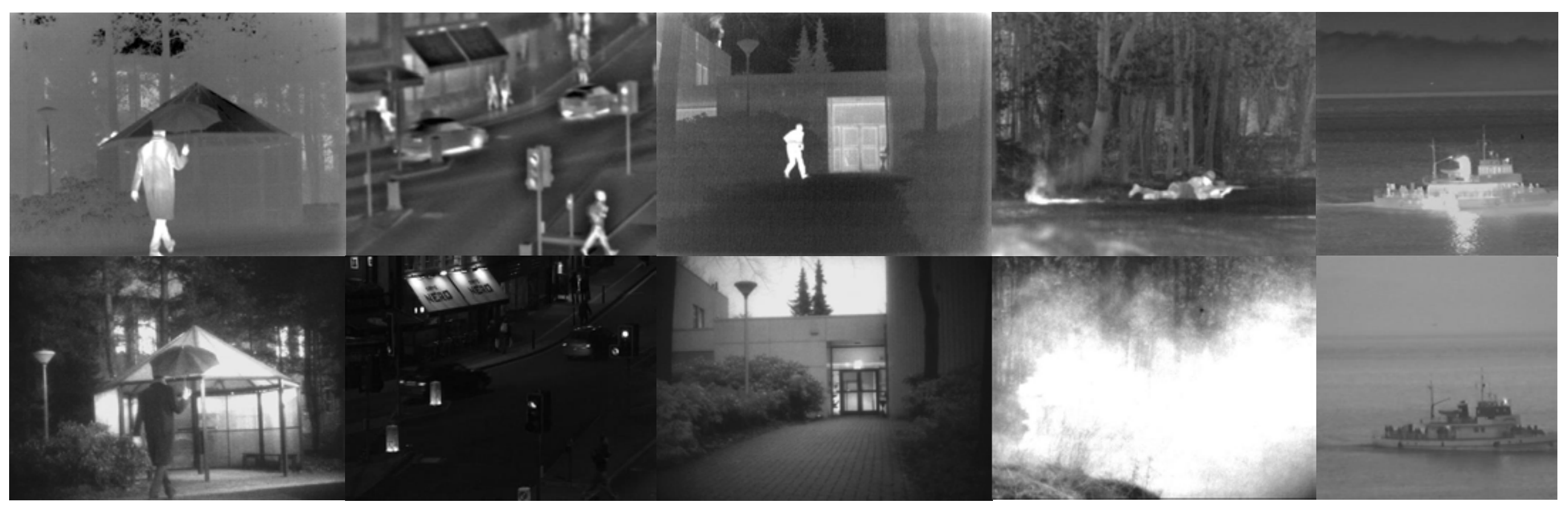}
\caption{Five pairs of source images. The top row contains infrared images, and the second row contains visible images.}
\label{fig:testing}
\end{figure}

\subsection{Experimental Setting}

The test data is collected from \cite{tno2018}, it contains 21 pairs infrared and visible images. A sample of these infrared and visible images is shown in Figure \ref{fig:testing}. The code of MDLatLRR and the test data are available at \cite{li2019codes}\footnote{The test data path is ``/IV\_images/''. For more results on additional test data (new test data containing 20 pairs of images), please refer to supplementary materials}. 

\begin{figure*}[ht]
\centering
\includegraphics[width=0.95\linewidth]{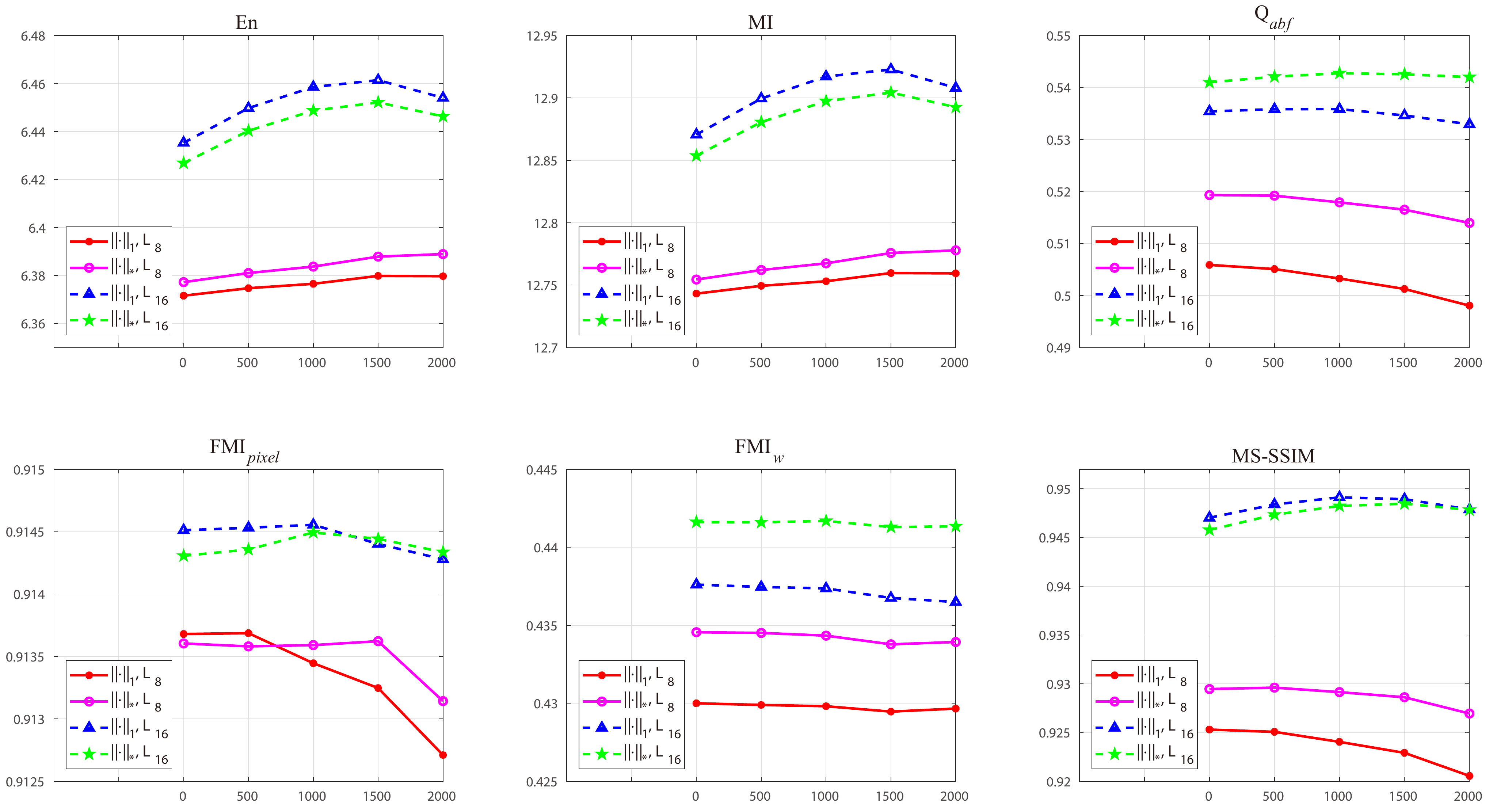}
\caption{The fusion performance of different projection matrices. The number of training image patches is 2000. The horizontal axis indicates the number of smooth patches in all training patches. The vertical axis denotes the average values of the six quality metrics computed on all test images. $||\cdot||_1$ and $||\cdot||_*$ indicate $l_1$-norm and nuclear norm, respectively. $L_8$ and $L_{16}$ denote projection matrices trained by different image patch size ($8\times 8$ and $16\times 16$).}
\label{fig:learning}
\end{figure*}

\begin{figure}[ht]
\centering
\includegraphics[width=0.85\linewidth]{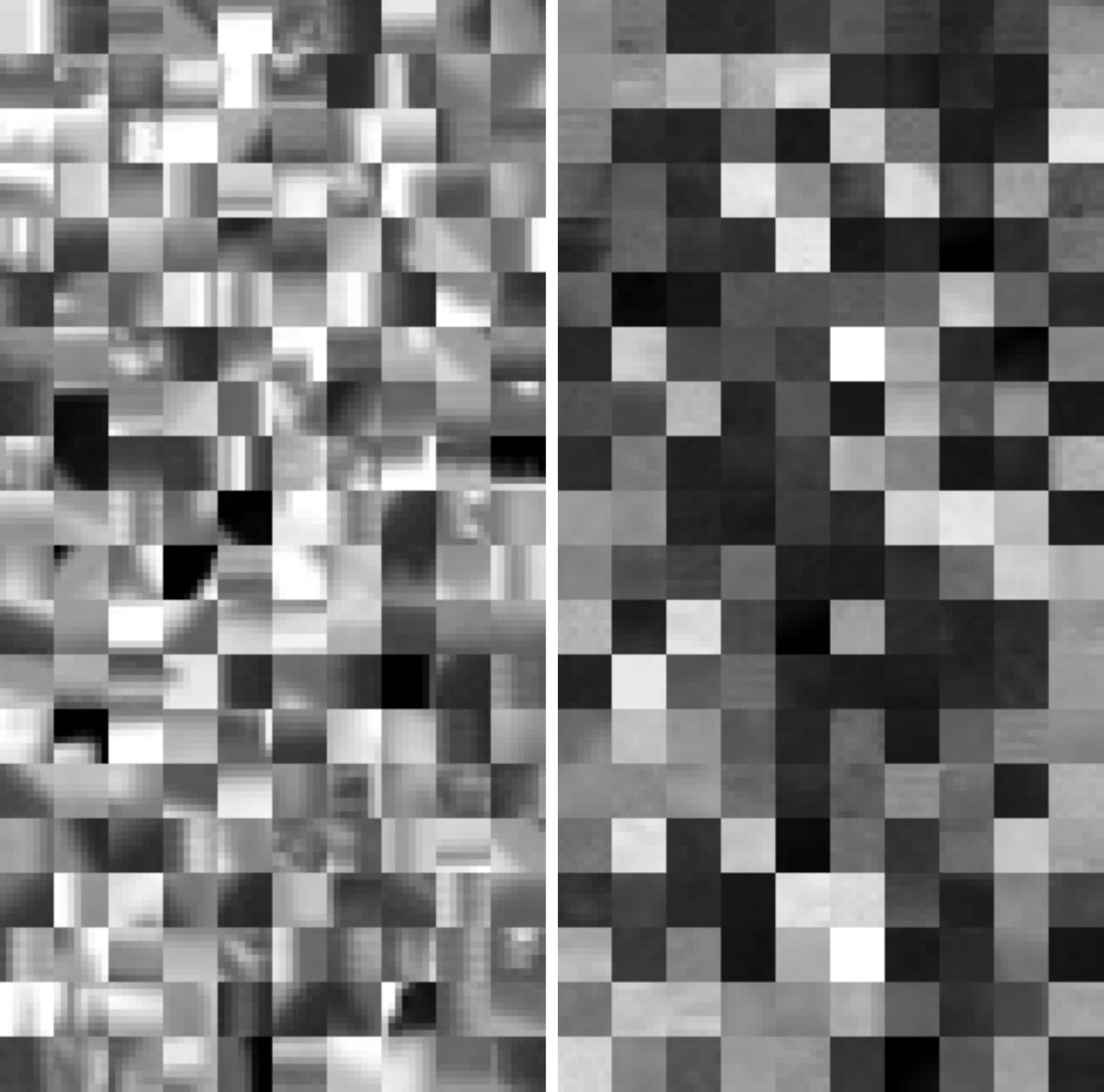}
\caption{Training image patches selection ($8\times 8$). Left part is 200 $detail$ image patches and right part is 200 $smooth$ patches, which are all randomly choosen from categorized image patches.}
\label{fig:l8de}
\end{figure}

\begin{figure}[ht]
\centering
\includegraphics[width=0.85\linewidth]{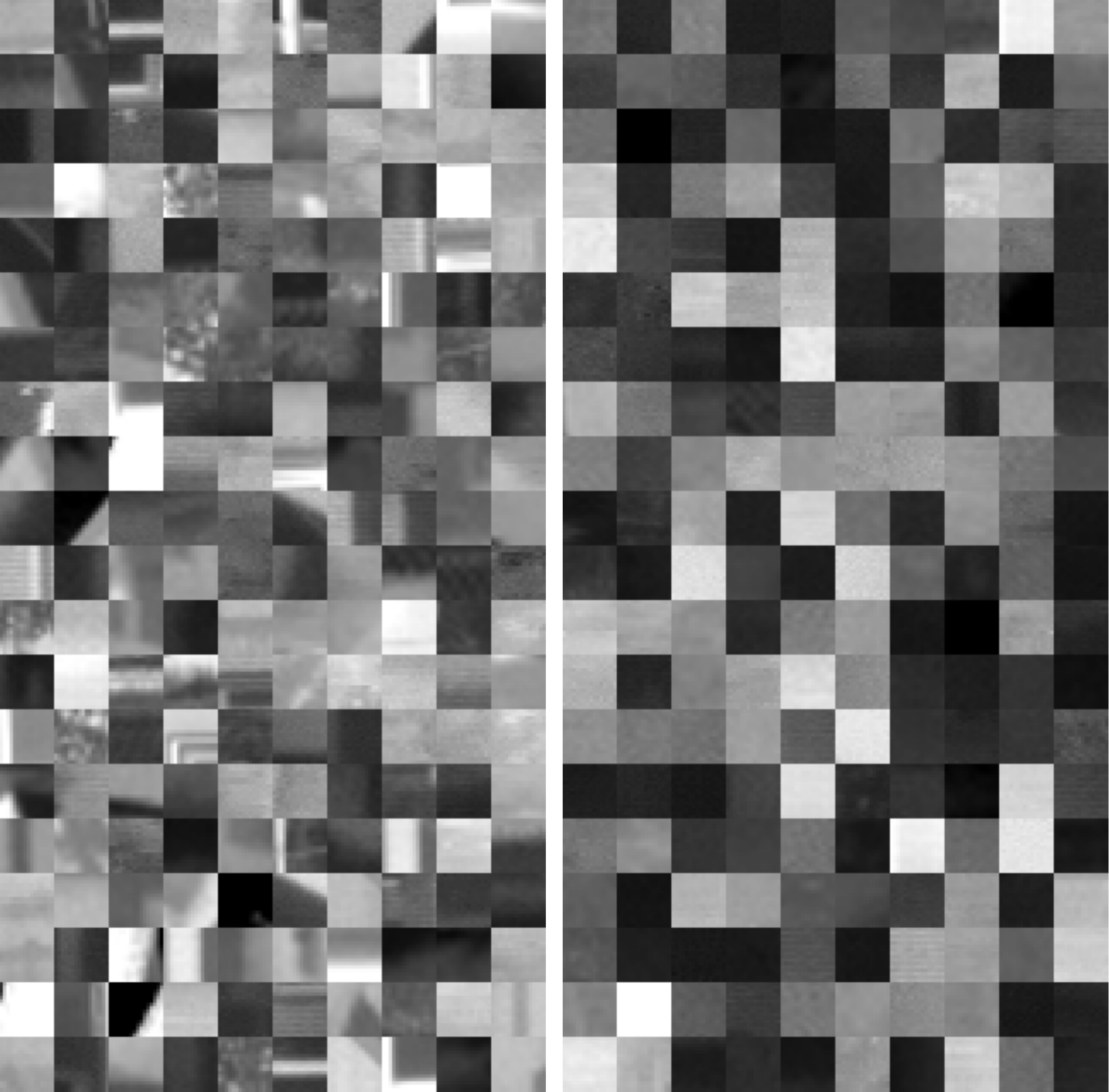}
\caption{Training image patches selection ($16\times 16$). Left part is 200 $detail$ image patches and right part is 200 $smooth$ patches, which are all randomly choosen from categorized image patches.}
\label{fig:l16de}
\end{figure}

\begin{figure*}[ht]
\centering
\includegraphics[width=0.95\linewidth]{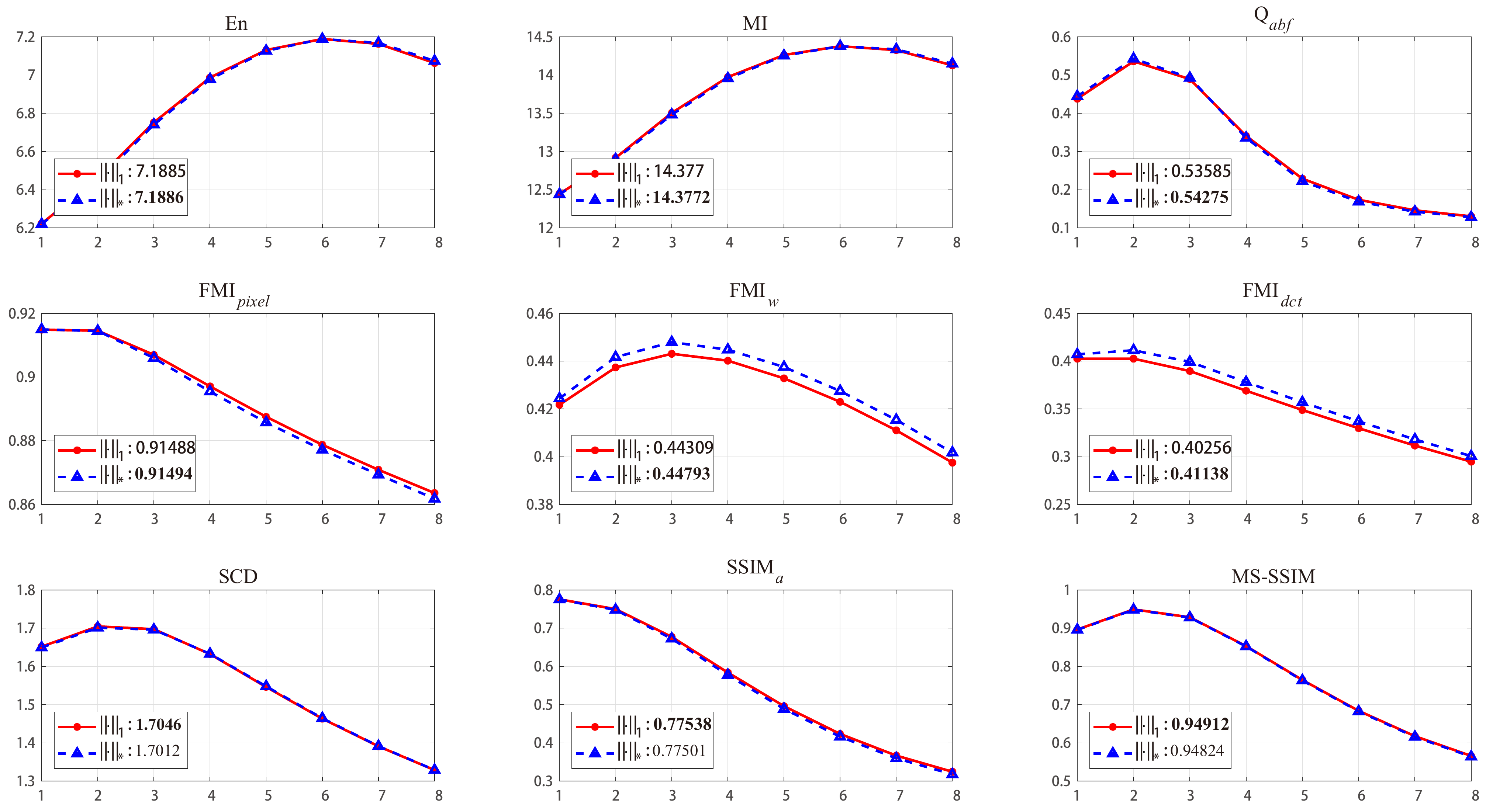}
\caption{Different decomposition levels and different norms. The average values of nine metrics($En$, $MI$, $Q_{abf}$, $FMI_{pixel}$, $FMI_w$, $FMI_{dct}$, $SCD$, $SSIM_a$ and MS-SSIM) for all fused images obtained by MDLatLRR using multi-level(8 levels, horizontal axis) and different norms. $||\cdot||_1$ denotes $l_1$-norm, and $||\cdot||_*$ indicates nuclear-norm.}
\label{fig:multi-levels}
\end{figure*}

\begin{figure*}[ht]
\centering
\includegraphics[width=0.95\linewidth]{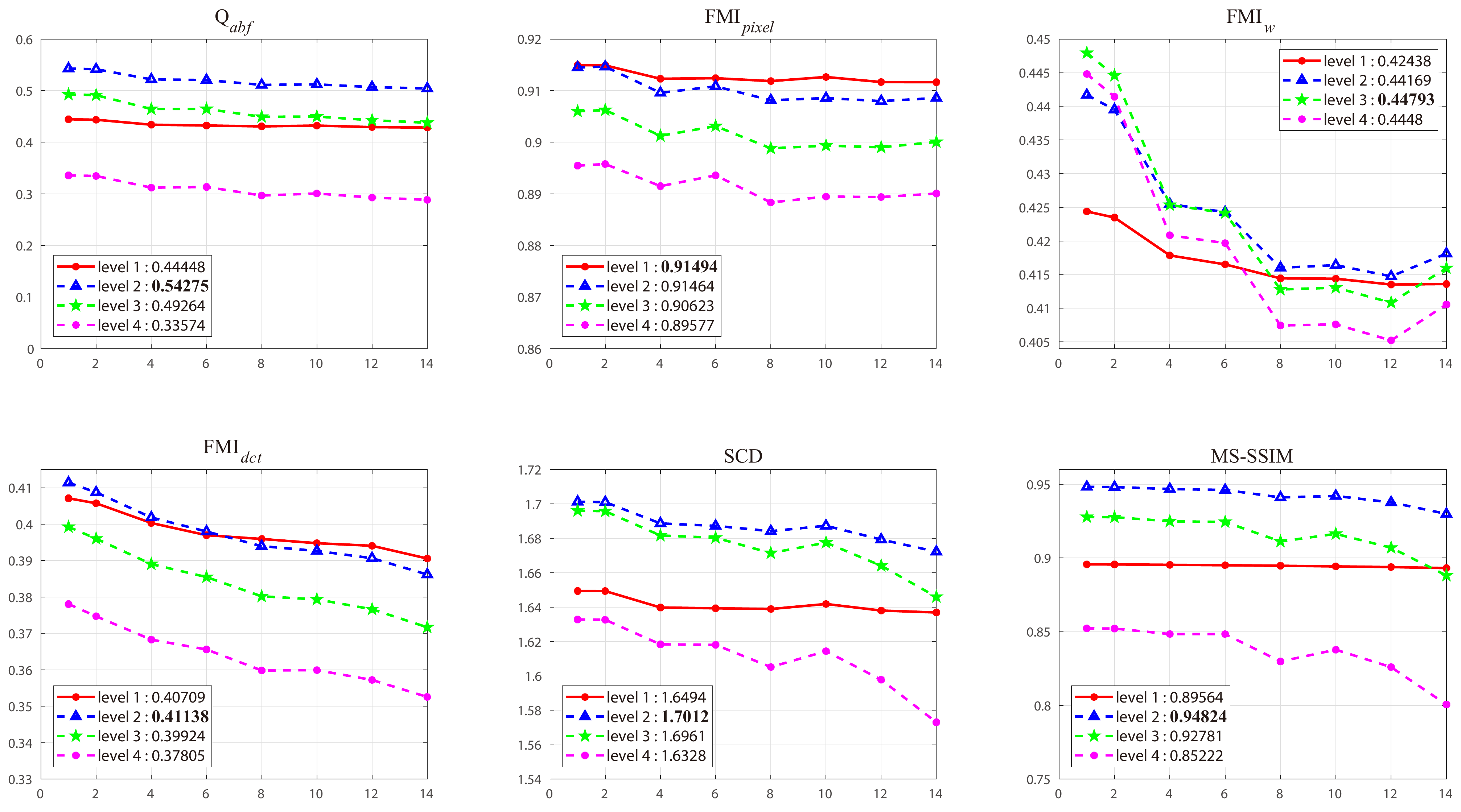}
\caption{The sliding window technique with different strides ($s=1, 2, 4, 6, 8, 10, 12, 14$). We present the average values of six metrics($Q_{abf}$, $FMI_{pixel}$, $FMI_w$, $FMI_{dct}$, $SCD$ and MS-SSIM) for all fused images obtained by MDLatLRR with different stride(8 strides, horizontal axis) and different levels(1 to 4).}
\label{fig:multi-strides}
\end{figure*}


Firstly, the sliding window technique is used to divide the source images into image patches, using the stride set to 1. The window size depends on the size of $L$ ($8\times8, 16\times16$).

In our MDLatLRR-based fusion method, $w_{b1}=w_{b2}=0.5$ for the fusion strategy of the base part, since the base parts contain more common features and redundant information.

For comparison, thirteen recent and classical fusion methods are chosen to conduct the same experiment. They include: cross bilateral filter method(CBF) \cite{kumar2015image}, discrete cosine harmonic wavelet transform method(DCHWT) \cite{kumar2013multifocus}, joint sparse representation-based method(JSR) \cite{zhang2013dictionary}, the JSR model with saliency detection method(JSRSD) \cite{liu2017infrared}, the gradient transfer fusion method(GTF) \cite{ma2016infrared}, weighted least square optimization-based method(WLS) \cite{ma2017infrared}, convolutional sparse representation(ConvSR) \cite{liu2016image}, VGG-19 and multi-layer fusion strategy-based method(VggML) \cite{li2018infrared}, ResNet50 and zero-phase component analysis fusion framework (ResNet-ZCA\footnote{The deep features were extracted by 'Conv5'.})\cite{li2019infrared}, a CNN-based method(DeepFuse) \cite{prabhakar2017deepfuse}, GAN-based fusion algorithm(FusionGAN) \cite{ma2019fusiongan}, dense block based fusion method(DenseFuse\footnote{The fusion strategy is 'addition' and $\lambda = 1e2$.}) \cite{li2018densefuse} and an end-to-end fusion network (IFCNN) \cite{zhang2020ifcnn}.

For the purpose of a quantitative comparison between our method and the other existing fusion methods, twelve quality metrics are utilized. These are: entropy($En$) \cite{roberts2008assessment}; mutual information($MI$) \cite{hossny2008comments} indicates how much features are preserved in fused images; the standard deviation ($SD$) \cite{rao1997fibre}; $Q_{abf}$ \cite{xydeas2000objective} reflects the quality of visual information obtained from the fusion of the input images; $FMI_{p}$, $FMI_w$ and $FMI_{dct}$ \cite{haghighat2014fast} calculate fast mutual information ($FMI$), discrete cosine and wavelet features, respectively; $SCD$ \cite{aslantas2015new} calculates image quality metric value based on the sum of the correlations of differences; a modified structural similarity $SSIM_a$ \cite{li2018infrared}; MS-SSIM \cite{ma2015perceptual} calculates a modified structural similarity which only focuses on structural information; visual information fidelity($VIF$) \cite{sheikh2006image}; and edge preservation index ($EPI$) \cite{sattar1997image}. The fusion result is better with the increasing magnitude of these metric values.

All the experiments are implemented in MATLAB R2017b on 2.8 GHz Intel(R) Core(TM) i5-8400 CPU with 16 GB RAM.

\begin{figure*}[ht]
\centering
\includegraphics[width=0.9\linewidth]{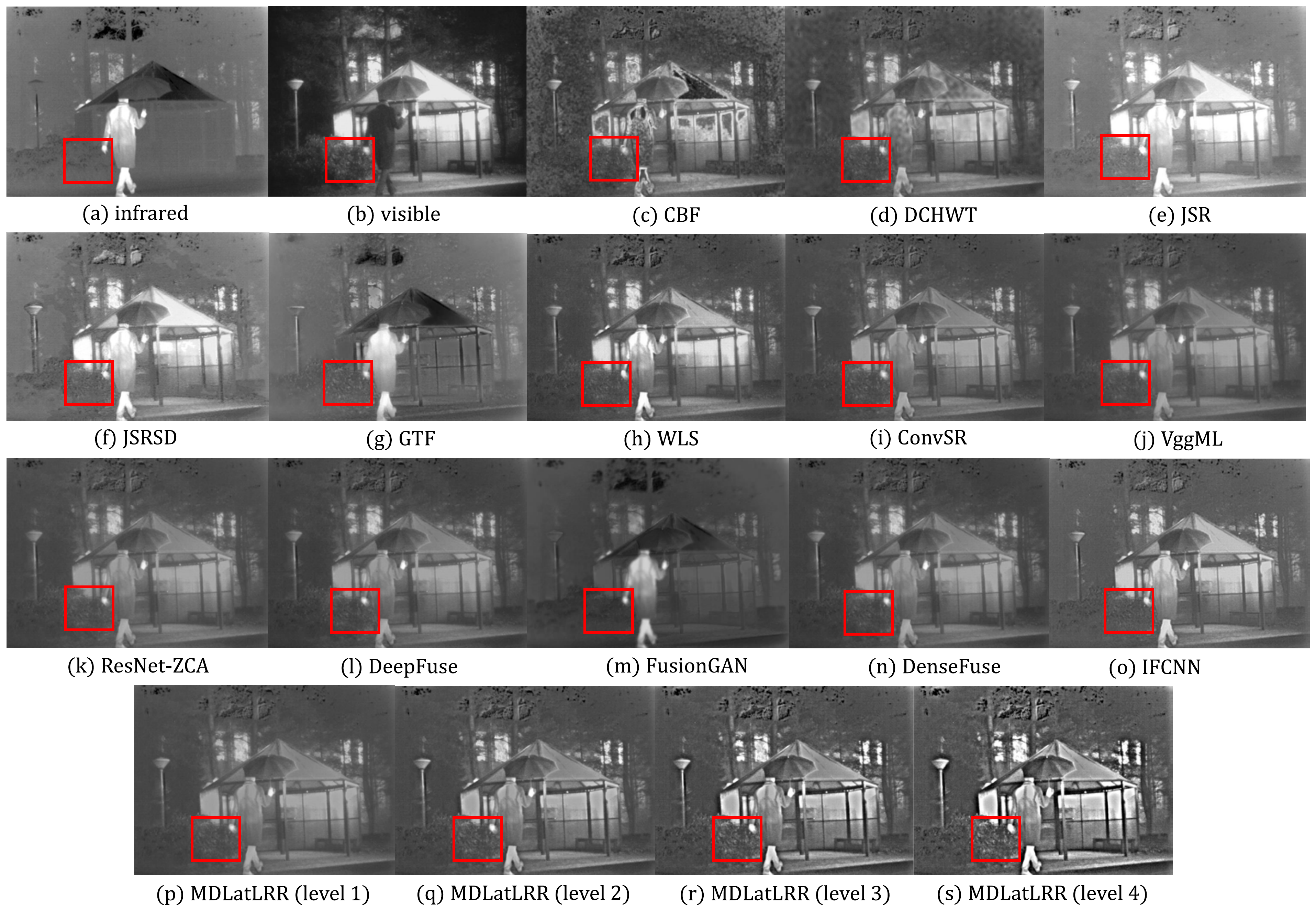}
\caption{ Experiment on ``umbrella'' images. (a) Infrared image; (b) Visible image; (c) CBF; (d) DCHWT; (e) JSR; (f) JSRSD; (g) GTF; (h) WLS; (i) ConvSR; (j) VggML; (k) ResNet-ZCA; (l) DeepFuse; (m) FusionGAN; (n) DenseFuse; (o) IFCNN; (p-s) MDLatLRR ($L_{16}$) with levels 1 to 4.}
\label{fig:results-umbrella}
\end{figure*}

\begin{figure*}[!ht]
\centering
\includegraphics[width=0.85\linewidth]{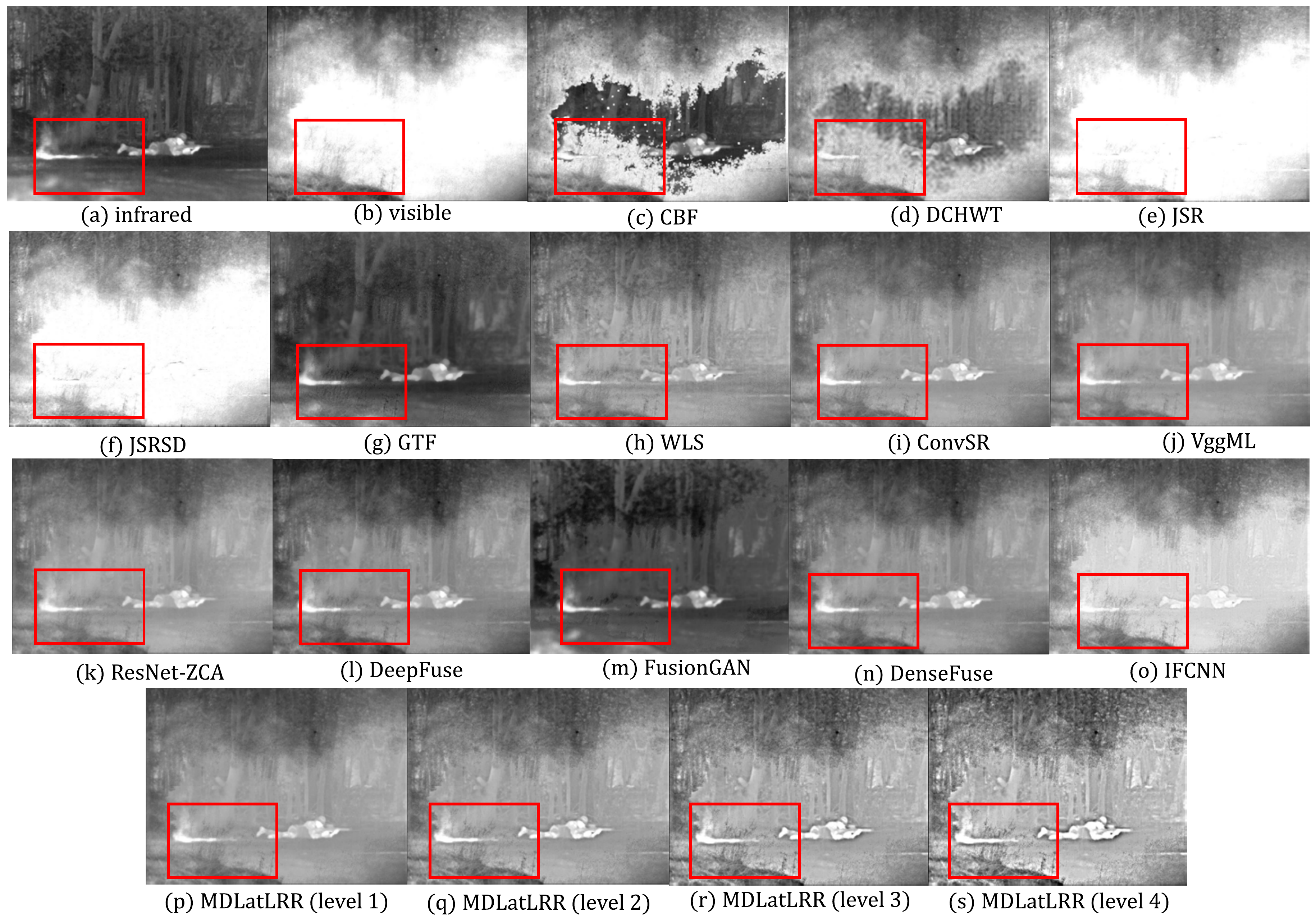}
\caption{ Experiment on ``soldier'' images.  (a) Infrared image; (b) Visible image; (c) CBF; (d) DCHWT; (e) JSR; (f) JSRSD; (g) GTF; (h) WLS; (i) ConvSR; (j) VggML; (k) ResNet-ZCA; (l) DeepFuse; (m) FusionGAN; (n) DenseFuse; (o) IFCNN; (p-s) MDLatLRR ($L_{16}$) with levels 1 to 4.}
\label{fig:results-soldier}
\end{figure*}

\begin{figure*}[!ht]
\centering
\includegraphics[width=0.85\linewidth]{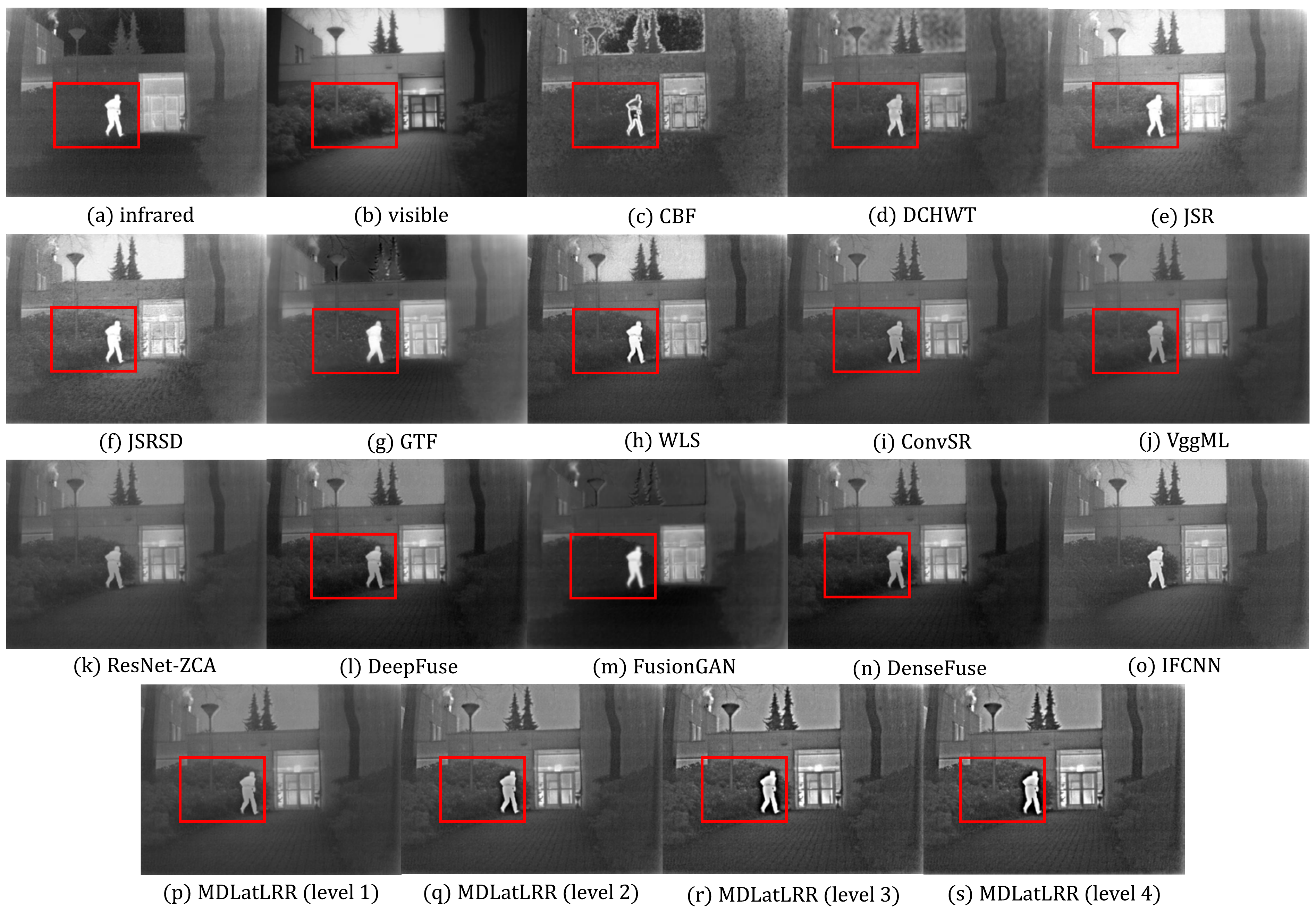}
\caption{ Experiment on ``man'' images. (a) Infrared image; (b) Visible image; (c) CBF; (d) DCHWT; (e) JSR; (f) JSRSD; (g) GTF; (h) WLS; (i) ConvSR; (j) VggML; (k) ResNet-ZCA; (l) DeepFuse; (m) FusionGAN; (n) DenseFuse; (o) IFCNN; (p-s) MDLatLRR ($L_{16}$) with levels 1 to 4.}
\label{fig:results-man}
\end{figure*}

\subsection{Ablation Study}
\label{ablation}

\subsubsection{\textbf{Projection Matrix $L$}}

In this section, we discuss how to choose the training image patches when learning the projection matrix $L$. 

As mentioned in Section \ref{learn-l}, the sliding window technique is used to divide training images into image patches. If the window size is $8\times 8$, 40654 image patches are obtained. For $16\times 16$, we will get 10090 image patches. These image patches are classified into two categories ($detail$, $smooth$) by standard deviation ($SD$) \cite{rao1997fibre}. The $SD$ is defined as,
\begin{eqnarray}\label{equ:sd}
  	SD(P)=\sqrt{\sum_{i=1}^{H}\sum_{j=1}^{W}(P(i,j)-\mu)^2}
\end{eqnarray}
where $P(i,j)$ is the pixel value of image patch, $H$ and $W$ denote the patch size ($H\times W$). $\mu$ is the mean value of the input image patch. The larger $SD$ means image patch contains more detail information. In this experiment, the classification strategy ($C$) is defined as,
\begin{equation}\label{equ:5}
	C(P)=\left\{\begin{array}{ll}
		detail & \textrm{ $SD(P)>Th$ } \\
        smooth & \textrm{ otherwise }
	\end{array}\right.
\end{equation}
where $P$ indicates the input image patch, $Th$ is the threshold which determines the patch belongs to $detail$ or $smooth$. $Th$ is set to 0.5 in this experiment.

Following the classification strategy ($C$), for $8\times 8$ image patches, 2316 $detail$ patches and 38338 $smooth$ patches are obtained. For $16\times 16$, we have 2646 $detail$ patches and 7444 $smooth$ patches, respectively. The examples of these patches are shown in Figure \ref{fig:l8de} and Figure \ref{fig:l16de}, respectively.

We randomly choose 2000 image patches to learn the projection matrix $L$. Thus, the number of $detail$ and $smooth$ in 2000 training patches becomes an important issue. 

We set the number of $smooth$ to 0, 500, 1000, 1500, and 2000 to analyze the influence of the ratio $detail$ to $smooth$. Then, five group ratios are obtained which are [2000, 0], [1500, 500], [1000, 1000], [500, 1500], [0, 2000]. The first value is the $detail$ number and the second is the $smooth$ number. The projection matrices are learned by the above training patches and the decomposition level of MDLatLRR is temporarily set to 2. For each scale ($8\times 8$, $16\times 16$), five matrices are learned by LatLRR. Six quality metrics ($En$, $MI$, $Q_{abf}$, $FMI_{pixel}$, $FMI_w$ and MS-SSIM) are chosen to evaluate the performance for different projection matrices. The average values are shown in Figure \ref{fig:learning}. 

From Figure \ref{fig:learning}, when the size of image patch is $16\times 16$, the proposed method obtains better fusion performance no matter what the ratios are. The reason is that a larger image patch contains more useful information, which is beneficial for generating fused image. However, if the image patch size is $32\times 32$, the size of the projection matrix will become $1024\times 1024$, and the MDLatLRR fusion method will become time-consuming. 

Focusing on $L_{16}$ (with $l_1$-norm and nuclear norm), the proposed method obtains four best values ($Q_{abf}$, $FMI_{pixel}$, $FMI_w$ and MS-SSIM) when the ratio is [1000, 1000]. For $En$ and $MI$, the projection matrix $L$ still achieves comparable performance. 

The reason why 2000 $detail$ patches can not learn a better projection matrix is that both positive and negative samples should be available for learning. In our learning process, $detail$ patches are positive samples and $smooth$ patches are negative samples. Hence, in our experiments, the projection matrix $L$ is learned using 1000 $detail$ patches and 1000 $smooth$ patches with the size of image patch being $16\times 16$. $l_1$-norm and nuclear norm are used to fuse detail parts.

\subsubsection{\textbf{Decomposition Levels and Nuclear-norm}}

In this section, multi-level decompositions, $l_1$-norm and nuclear-norm are utilized to make comparisons. When $l_1$-norm is used in fusion, Eq \ref{equ:nunorm} is changed to Eq \ref{equ:l1norm},
\begin{eqnarray}\label{equ:l1norm}
\begin{split}
  	&w_{dk}^{i,j}=\frac{\hat{w_{dk}^{i,j}}}{\sum_{q=1}^K\hat{w_{dq}^{i,j}}},\quad \hat{w_{dk}^{i,j}}=||V_{dk}^{i,j}||_1
\end{split}
\end{eqnarray}
\noindent where $||\cdot||_1$ indicates the $l_1$-norm.

The average values of the nine quality metrics are shown in Figure \ref{fig:multi-levels}. The annotation contains the maximum values in each different situation and the best values are marked in \textbf{bold}.

In Figure \ref{fig:multi-levels}, the levels between 1 to 4 obtain almost best values, but the increasing the number of decomposition levels(more than four levels) would cause fusion performance degradation. With the growing number of decomposition levels, the detail parts will contain more base information (e.g. luminance, contour). However, the fusion strategy for detail parts is not equipped to fuse them. So, in image fusion tasks, the decomposition levels should be chosen carefully. From Figure \ref{fig:multi-levels}, when the number of decomposition levels is set to 1 to 4, our proposed fusion method achieves seven best values (except $En$ and $MI$). Even for $En$ and $MI$, the results are still comparable when the level is set to 3 and 4. The reason is that the source images are successfully divided into the detail parts and base part by our MDLatLRR with level 1 to 4, and this is beneficial for image fusion.

On the other hand, the nuclear norm achieves six best values in all nine quality metrics. For $SCD$, $SSIM_a$ and MS-SSIM, the nuclear norm still reports comparable values. These results indicate the fusion strategy with the nuclear-norm achieves better performance than other option ($l_1$-norm). Recall, the nuclear-norm, which calculates the sum of singular values of a matrix, represents the structural information of an image patch. 

In our nuclear-norm based fusion strategy, when the detail patch contains more texture and structural information, which means larger values of nuclear-norm, this type of patch will be endowed larger weight in the fusion process. Thus, the nuclear-norm based strategy leads to better fusion performance.

Based on this analysis, in our MDLatLRR fusion framework, the projection matrix is $L_{16}$, the decomposition level varies between 1 to 4 (means $r=1,2,3,4$) and nuclear-norm is used in the detail parts fusion strategy.

\subsubsection{\textbf{The Stride of Sliding Window Technique}}

In the sliding window technique, different strides result in different fusion performance. In our experiment, the stride ($s$) is set to $\{1, 2, 4, 6, 8, 10, 12, 14\}$ to analyze the influence.

As discussed before, the image patch size is $16\times 16$, the projection matrix is $L_{16}$ and the decomposition level is set to $\{1, 2, 3, 4\}$. We choose six quality metrices ($Q_{abf}$, $FMI_{pixel}$, $FMI_w$, $FMI_{dct}$, $SCD$ and MS-SSIM) to evaluate the fusion performance with different strides in the sliding window technique. The average metric values of all test images are shown in Figure \ref{fig:multi-strides}.

In Figure \ref{fig:multi-strides}, when the decomposition level is set to 1, the quality metrics (except $FMI_{w}$ and $FMI_{dct}$) values are quite stable. The reason is that the detail information is not fully extracted by MDLatLRR with the shallow decomposition level (such as level 1), while the base parts contain more source image information. Thus, in this case, the stride has less influence. 

However, with the increase in the number of decomposition levels, the detail parts preserve more texture and structural information, and our proposed method achieves better fusion performance. Thus, the stride of the sliding window should be chosen carefully when the decomposition level deepens. 

In all six metrices, increasing the stride leads to fusion performance degradation no matter what the decomposition levels are (\emph{level-2}, \emph{level-3} and \emph{level-4}). When the stride is large (more than 2), less feature information is processed by the projection matrix and the detail parts contain much less detail information. We can observe that more usefull information is extracted by a small stride, and the overall fusion performance is improved. However, the small stride also makes our fusion method become time-consuming. Thus, the choice of stride should be optimized for different tasks.

In our experiment, the stride is set to 1 to obtain the best fusion performance. When the proposed method is applied to other computer vision tasks (such as RGBT object tracking), the time efficiency is an important factor, it is also acceptable that the stride is set to 4. This experiment is discussed in Section \ref{tracking}.

\begin{table*}[ht]
\centering
\scriptsize
\caption{\label{tab:compared}The average values of ten quality metrics for all source images.}
\resizebox{\linewidth}{!}{
\begin{tabular}{|c|c|c|c|c|c|c|c|c|c|c|c|}
\hline 
\multicolumn{2}{|c|}{}	& En & MI  & $SD$ & $Q_{abf}$   & $FMI_{p}$  & $FMI_w$ & $SSIM_a$ & MS-SSIM & VIF & EPI \\
\hline
\multicolumn{2}{|c|}{CBF \cite{kumar2015image}}			&\emph{\color{red}{6.85749}} 	&\emph{\color{red}{13.71498}} 	&76.82410 	&0.43961 	&0.87203 	&0.32350 	&0.59957 	&0.70879 	&0.71849 	&0.57240 \\ 
\hline
\multicolumn{2}{|c|}{DCHWT \cite{kumar2013multifocus}}			&6.56777 	&13.13553 	&64.97891	&0.46592 	&0.91377 	&0.40147 	&0.73132 	&0.84326 	&0.50560 	&0.70181 \\ 
\hline
\multicolumn{2}{|c|}{JSR\cite{zhang2013dictionary}}			&6.72263 	&12.72654 	&74.10783	&0.32306 	&0.88463 	&0.18506 	&0.54073 	&0.75523 	&0.63845 	&0.59644 \\ 
\hline
\multicolumn{2}{|c|}{JSRSD\cite{liu2017infrared}}			&6.72057 	&13.38575 	&\emph{\color{red}{79.19536}}	&0.32281 	&0.86429 	&0.18498 	&0.54127 	&0.75517 	&0.67071 	&0.47473 \\ 
\hline
\multicolumn{2}{|c|}{GTF\cite{ma2016infrared}}			&6.63433 	&13.26865 	&67.54361	&0.41037 	&0.90207 	&0.41038 	&0.70016 	&0.80844 	&0.41687 	&0.67011 \\ 
\hline
\multicolumn{2}{|c|}{WLS\cite{ma2017infrared}}			&6.64071 	&13.28143 	&70.58894	&0.50077 	&0.89788 	&0.37662 	&0.72360 	&0.93349 	&0.72874 	&0.67837 \\ 
\hline
\multicolumn{2}{|c|}{ConvSR\cite{liu2016image}}		&6.25869 	&12.51737 	&50.74372	&\emph{\color{red}{0.53485}} 	&\textbf{0.91691} 	&0.34640 	&0.75335 	&0.90281 	&0.39218 	&0.71130 \\ 
\hline
\multicolumn{2}{|c|}{VggML\cite{li2018infrared}}			&6.18260 	&12.36521 	&48.15779	&0.36818 	&0.91070 	&0.41684 	&\emph{\color{red}{0.77799}} 	&0.87478 	&0.29509 	&0.77923 \\ 
\hline
\multicolumn{2}{|c|}{ResNet-ZCA\cite{li2019infrared}}	&6.19527 	&12.39054 	&48.71803	&0.35098 	&0.90921 	&0.41689 	&\textbf{0.77825} 	&0.87324 	&0.29249 	&\textbf{0.78366} \\ 
\hline
\multicolumn{2}{|c|}{DeepFuse\cite{prabhakar2017deepfuse}}		&6.69935 	&13.39869 	&68.79312	&0.43797 	&0.90470 	&0.42477 	&0.72882 	&\emph{\color{red}{0.93353}} 	&0.65773 	&0.73795 \\ 
\hline
\multicolumn{2}{|c|}{FusionGAN\cite{ma2019fusiongan}}		&6.36285 	&12.72570 	&67.57282	&0.21834 	&0.89061 	&0.37083 	&0.65384 	&0.73182 	&0.45355 	&0.68470 \\ 
\hline
\multicolumn{2}{|c|}{DenseFuse\cite{li2018densefuse}}		&6.67158 	&13.34317 	&54.35752	&0.44009 	&0.90847 	&0.42767 	&0.73150 	&0.92896 	&0.64576 	&0.73321 \\ 
\hline
\multicolumn{2}{|c|}{IFCNN\cite{zhang2020ifcnn}}			&6.59545 	&13.19090 	&66.87578	&0.50329 	&0.90092 	&0.40166 	&0.73186 	&0.90527 	&0.59029 	&0.73767 \\ 
\hline
\multirow{4}*{MDLatLRR} &
 level-1	&6.21986 	&12.43972 	&49.65098	&0.44448 	&\emph{\color{red}{0.91494}} 	&0.42438 	&0.77501 	&0.89564 	&0.34773 	&\emph{\color{red}{0.76350}} \\ 
&level-2	&6.44870 	&12.89740	&57.69733	&\textbf{0.54275} 	&0.91449 	&0.44169 	&0.74783 	&\textbf{0.94824} 	&0.71472 	&0.71142 \\ 
&level-3	&6.74064 	&13.48128 	&69.91578	&0.49264 	&0.90599 	&\textbf{0.44793} 	&0.67248 	&0.92781 	&\emph{\color{red}{1.39009}} 	&0.66903 \\ 
&level-4	&\textbf{6.97774} 	&\textbf{13.95548}	&\textbf{83.77407} 	&0.33574 	&0.89546 	&\emph{\color{red}{0.44480}} 	&0.57736 	&0.85222 	&\textbf{2.27705} 	&0.63077 \\ 
\hline
\end{tabular}}
\end{table*}

\begin{figure*}[ht]
\centering
\includegraphics[width=\linewidth]{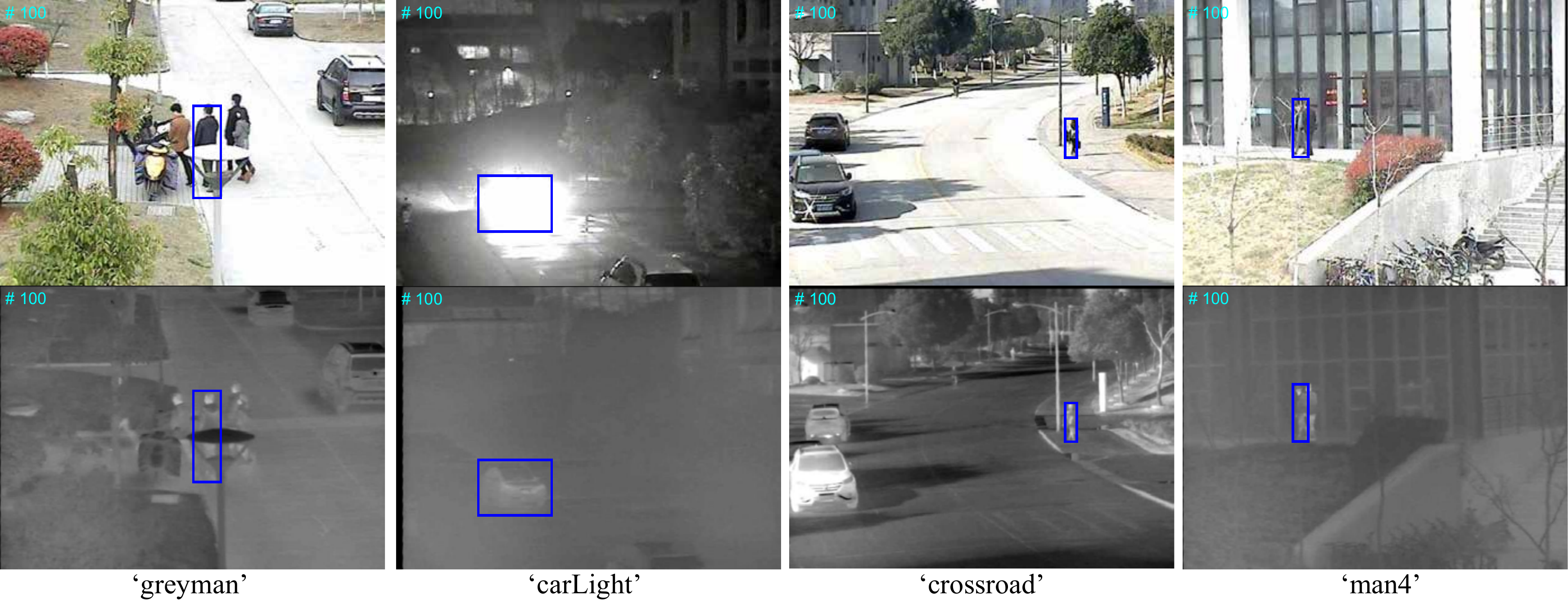}
\caption{Examples of RGBT frames in the VOT-RGBT2019 benchmark. Each cloumn denotes a pair of frames which come from a video. First row shows RGB frames and the infrared frames are shown in the second row. The groundtruth of each frame is indicated by the blue box.}
\label{fig:frames-rgbt}
\end{figure*}

\subsection{Subjective Evaluation}
\label{subjective}

The fused results for three pairs of source images (``umbrella'', ``soldier'' and ``man'')\footnote {More experiments are presented in our supplementary materials.} are shown in Figure \ref{fig:results-umbrella}-\ref{fig:results-man}. These results are obtained by thirteen existing fusion methods and our algorithm (MDLatLRR), which uses $L_{16}$, nuclear-norm and decomposition levels (1 to 4). For sliding window technique, the stride is set to 1.

In Figure \ref{fig:results-umbrella}-\ref{fig:results-man}, the fused images obtained by CBF and DCHWT contain more artefacts and noise, and their salient features are not clear. The fused images obtained by JSR, JSRSD and GTF contain many artefacts, manifesting in ringing around the salient features. 

In contrast, the fused images obtained by WLS, ConvSR, VggML, ResNet-ZCA, DeepFuse, FusionGAN, DenseFuse, IFCNN and the proposed method capture more salient features and the fused images have a better visual quality.

Specifically for the bushes(red boxes) in Figure \ref{fig:results-umbrella}-\ref{fig:results-man}, comparing with the state-of-the-art fusion methods, MDLatLRR preserves more detailed information.

On the other hand, with the increasing number of decomposition levels(between 1 to 4), the salient features are better enhanced by our fusion methods and the detail information becomes more clear, as shown in Figure \ref{fig:results-umbrella}-\ref{fig:results-man}(p-s).

From these figures, our fusion method preserves more salient features from the infrared image and more detail information from the visible image in the subjective evaluation.

\subsection{Objective Evaluation}
\label{objective}

In order to demonstrate  the attractive properties of our fusion method, ten quality metrics are used to compare the fusion performance of thirteen existing fusion methods and our algorithm.

The average values of the ten quality metrics for the set of test images are presented in Table \ref{tab:compared}. The best and the second-best values are marked  in \textbf{bold} and \emph{\color{red}{red italic}}, respectively.

In Table \ref{tab:compared}, the proposed method achieves the best results in seven cases ($En$, $MI$, $SD$, $Q_{abf}$, $FMI_w$, MS-SSIM and $VIF$) and second-best values in four cases($FMI_{p}$, $FMI_w$, $VIF$ and $EPI$). Although in $SSIM_a$, our method dose not achieve the best value, it obtains a comparable result. These values indicate that the fused images obtained by our method are more natural and contain fewer artefacts. The objective evaluation shows that our fusion method delivers better fusion performance than the existing methods.

\begin{figure*}[ht]
\centering
\includegraphics[width=\linewidth]{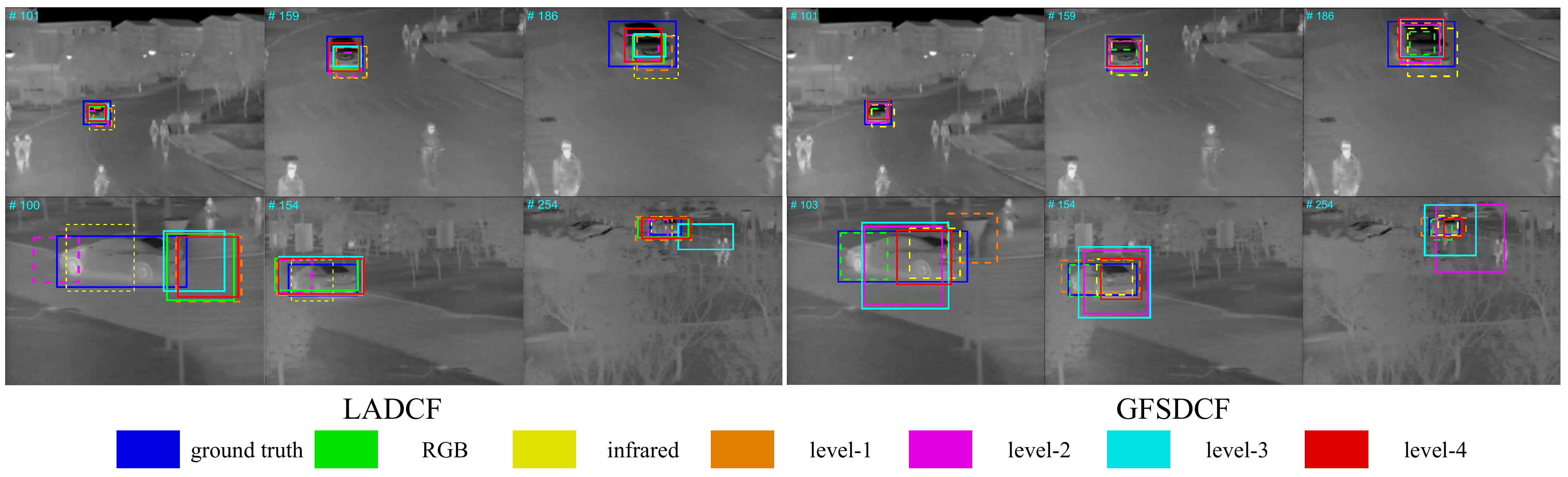}
\caption{The tracking results on the VOT-RGBT2019 benchmark. The frames in first row and second row are selected from `car10' and `car41', respectively. First three columns are the results of LADCF. The last three columns are the tracking results of GFSDCF. The \emph{RGB} and \emph{infrared} denote the case when the input of the trackers is just one modality (RGB or infrared). The \emph{level-1} to \emph{level-4} present the cases when the input of the trackers are the fused frames generated by MDLatLRR.}
\label{fig:rgbt-all}
\end{figure*}

Specifically for the quality metrics of $EN$ and $MI$, we note that CBF achieved the second-best values. Its fused images contain more noise and artefacts, as shown in Figure \ref{fig:results-umbrella}-\ref{fig:results-man}(c). However, our fusion method achieves the best values on $EN$ and $MI$, thanks to the enhancement of the salient features, instead of amplifying noise and generating artefacts. Thus our fusion method also exhibits a feature enhancement capability.

\subsection{Fusion Method for Visual Object Tracking(RGB-T)}
\label{tracking}


As described in VOT2019\cite{vot2019rgbt}, the visual objecting tracking challenges address short-term or long-term, causal and model-free tracking\cite{xu2019learning}\cite{xu2019joint}\cite{xu2019an}. With the wide application of multiple sensors in computer vision, it is important to carefully process multi-modal data. Thus, in VOT2019, the committee introduced two new sub-challenges (VOT-RGBT and VOT-RGBD). The VOT-RGBT sub-challenge focuses on short-term tracking. It contains two modalities (RGB and thermal infrared). The RGBT tracking problem is ideally suited to take  advantage of our fusion method to improve the tracking performance.

In this experiment, we apply our fusion method to the RGBT visual object tracking task to verify the efficiency of our proposed fusion method in other computer vision task. The VOT-RGBT benchmark is available at\cite{vot2019rgbt} which contains 60 video sequences. The examples of these video frames are shown in Figure \ref{fig:frames-rgbt}.

\begin{table}[!ht]
\scriptsize
\centering
\caption{\label{tab:tracking-ladcf}Tracking results using LADCF and MDLatLRR on the VOT-RGBT2019 dataset. MDLatLRR is used to fuse RGB and infrared frames.}
\resizebox{\linewidth}{!}{
\begin{tabular}{|c|c|c|c|c|}
\hline
 & \emph{Type} &EAO & Accuracy & Failures \\
\hline
\multirow{2}*{{LADCF\cite{xu2019learning}}} & 
   \emph{RGB}		& 0.2518 & 0.5406 & 41.2745 \\
\cline{2-5}
~& \emph{infrared}	& 0.1613 & 0.5109 & 57.6843  \\
\hline
\multirow{4}*{\makecell[cc]{\textbf{LADCF}\\+\\\textbf{MDLatLRR}}} &
   \emph{level-1} 	& 0.2553 & \emph{\color{red}{0.5650}} & 40.4008 \\
\cline{2-5}
~& \emph{level-2} 	& \textbf{0.2633} & \textbf{0.5740} & 42.4033 \\
\cline{2-5}
~& \emph{level-3}  	& 0.2608 & 0.5604 & \emph{\color{red}{38.8786}} \\
\cline{2-5}
~& \emph{level-4} 	& \emph{\color{red}{0.2618}} & 0.5331 & \textbf{37.8026} \\
\hline
\end{tabular}}
\end{table}

Firstly, the RGB and infrared frames are fused by our MDLatLRR fusion method. The parameters of MDLatLRR are the same as in the fusion task except the stride ($s$) is set to 4. Then, the fused frame is fed into different trackers\footnote{The combination framework of our fusion method and trackers is shown in supplementary materials.}. To evaluate the effect of our fusion method on RGBT tracking, we choose two trackers (LADCF\cite{xu2019learning}, GFSDCF\cite{xu2019joint}) to evaluate the tracking performance on the VOT-RGBT2019 benchmark. 

The LADCF\cite{xu2019learning} obtained the best tracking performance on the public dataset of VOT-ST2018 sub-challenge. Based on LADCF, the GFSDCF\cite{xu2019joint} explored group feature selection strategies to reduce the redundancy of deep features. We use the VOT2018 version code of LADCF. The parameters of GFSDCF were learned on OTB-2015\cite{wu2015object} dataset. The codes of these trackers were provided by the original authors and could be downloaded from \cite{trwotrackers}.

As in VOT2019\cite{vot2019rgbt}, we choose three measures to analyze the tracking performance: accuracy, failures and expected average overlap (EAO). The accuracy indicates the average overlap between the predicted and ground truth bounding boxes during successful tracking periods. The failure measures the robustness of a tracker. The third primary measure, EAO, is an estimator of the average overlap a tracker manages to attain on a large collection with the same visual properties as the ground-truth. The larger the values of accuracy and EAO, the better tracking performance, and the lower the failure value the better robustness.

\begin{table}[!ht]
\scriptsize
\centering
\caption{\label{tab:tracking-gfsdcf}Tracking results using GFSDCF and MDLatLRR on the VOT-RGBT2019 dataset. MDLatLRR is used to fuse RGB and infrared frames.}
\resizebox{\linewidth}{!}{
\begin{tabular}{|c|c|c|c|c|}
\hline
 & \emph{Type} &EAO & Accuracy & Failures \\
\hline
\multirow{2}*{{GFSDCF\cite{xu2019joint}}} & 
   \emph{RGB}		& \emph{\color{red}{0.3104}} & 0.5832 & \emph{\color{red}{31.9099}} \\
\cline{2-5}
~& \emph{infrared}	& 0.1941 & 0.5608 & 55.5163 \\
\hline
\multirow{4}*{\makecell[cc]{\textbf{GFSDCF}\\+\\\textbf{MDLatLRR}}} &
   \emph{level-1} 	& 0.3032 & \textbf{0.6098} & 35.3073 \\
\cline{2-5}
~& \emph{level-2} 	& 0.3092 & 0.5916 & 32.1395 \\
\cline{2-5}
~& \emph{level-3}  	& 0.2963 & \emph{\color{red}{0.5989}} & 34.3755 \\
\cline{2-5}
~& \emph{level-4} 	& \textbf{0.3107} & 0.5824 & \textbf{29.0613} \\
\hline
\end{tabular}}
\end{table}

The evaluation measure values of LADCF and GFSDCF with our fusion method computed on VOT-RGBT are shown in Table \ref{tab:tracking-ladcf} and Table \ref{tab:tracking-gfsdcf}, respectively. The best values and the second-best values are marked in \textbf{bold} and \emph{\color{red}{red italic}}, respectively. We also choose several frames to demonstrate the tracking results\footnote{More tracking results are shown in supplementary materials.} on VOT-RGBT2019. These are shown in Figure \ref{fig:rgbt-all}. The ground truth is specified in the infrared-channel since this is the primary modality. The infrared frames are shown in Figure \ref{fig:rgbt-all}.

In Table \ref{tab:tracking-ladcf}, Table \ref{tab:tracking-gfsdcf} and Figure \ref{fig:rgbt-all}, the \emph{RGB} and \emph{infrared} denote that only one modality (RGB or infrared) is fed into the trackers. The \emph{level-1} to \emph{level-4} indicate that the RGB and infrared frames are firstly fused by our fusion method, then fed into the trackers. From these two tables, it is observed that with the pre-processing of our fusion method, both LADCF and GFSDCF trackers achieve better performance in three measures. Furthermore, when the decomposition level is increased, the trackers become more robust (lower values of Failures).

In VOT2019, the EAO is the primary measure. With the fused frame fed into the trackers, the value of EAO is increased. However, for GFSDCF, the increase is slight. The reason is that the fusion strategy in this experiment is pixel-level fusion which is particularly suited for extracting handcrafted features. Since LADCF has a greater proportion of handcrafted features, LADCF with \emph{level-2} decomposition obtains a higher EAO value. In GFSDCF, the deep features are more powerful, thus the influence of the fused frames is very limited (the EAO value obtained by the fused frame is close to the RGB frame).

In conclusion, even with the simplest way of feeding our fusion results into the trackers, the tracking performance is consistently improved. The fused images enhanced by our fusion method have a positive impact on this computer vision (RGBT tracking) application. Furthermore, if the tracker engages a greater proportion of deep features for the data representation, the fusion method should focus on feature-level fusion and decision-level fusion. Thus, in the future, we will focus on these directions to improve the RGBT tracker performance.

\section{Conclusions}
\label{conclusions}

In this paper, we proposed a novel multi-level decomposition method (MDLatLRR) for image decomposition. We also developed an MDLatLRR-based fusion framework for fusing infrared and visible images.  Firstly, a projection matrix $L$ is learned by LatLRR. This matrix is then utilized to extract detail parts and base parts of the input images at several representation levels. With MDLatLRR, multi-level salient features are extracted. The final fused image is reconstructed by adaptive fusion strategies designed specifically for dealing with the detail parts and the base parts, respectively. The MDLatLRR framework is general and can be used to provide an efficient decomposition approach for extracting multi-level features for an arbitrary number of input images. It can also be applied to other image processing fields with different projection matrix. The proposed method was evaluated both subjectively and objectively in a number of experiments. The experimental results demonstrate that the performance of the proposed method is superior to that of existing methods.

Furthermore, we also applied our proposed fusion method to a RGBT object tracking task. Even with the simplest combination of our fusion method and different trackers, the tracking performance was improved.

In the future, we will investigate the link between the proposed decomposition method and other decomposition methods. Other potential applications of the proposed decomposition method to other tasks of image processing will be considered.

{\small
\bibliographystyle{ieee}
\bibliography{bib-file}

\begin{thebibliography}{10}\itemsep=-1pt

\bibitem{aslantas2015new}
V.~Aslantas and E.~Bendes.
\newblock {A new image quality metric for image fusion: the sum of the
  correlations of differences}.
\newblock {\em Aeu-international Journal of electronics and communications},
  69(12):1890--1896, 2015.

\bibitem{ben2005multiscale}
A.~Ben~Hamza, Y.~He, H.~Krim, and A.~Willsky.
\newblock {A multiscale approach to pixel-level image fusion}.
\newblock {\em Integrated Computer-Aided Engineering}, 12(2):135--146, 2005.

\bibitem{borsoi2019super}
R.~A. Borsoi, T.~Imbiriba, and J.~C.~M. Bermudez.
\newblock {Super-resolution for hyperspectral and multispectral image fusion
  accounting for seasonal spectral variability}.
\newblock {\em IEEE Transactions on Image Processing}, 29:116--127, 2019.

\bibitem{chen2020infrared}
J.~Chen, X.~Li, L.~Luo, X.~Mei, and J.~Ma.
\newblock {Infrared and visible image fusion based on target-enhanced
  multiscale transform decomposition}.
\newblock {\em Information Sciences}, 508:64--78, 2020.

\bibitem{chen2019sparse}
Z.~Chen, X.-J. Wu, and J.~Kittler.
\newblock {A sparse regularized nuclear norm based matrix regression for face
  recognition with contiguous occlusion}.
\newblock {\em Pattern Recognition Letters}, 2019.

\bibitem{chen2019noise}
Z.~Chen, X.-J. Wu, H.-F. Yin, and J.~Kittler.
\newblock {Noise-Robust Dictionary Learning with Slack Block-Diagonal Structure
  for Face Recognition}.
\newblock {\em Pattern Recognition}, page 107118, 2019.

\bibitem{gao2017image}
R.~Gao, S.~A. Vorobyov, and H.~Zhao.
\newblock {Image fusion with cosparse analysis operator}.
\newblock {\em IEEE Signal Processing Letters}, 24(7):943--947, 2017.

\bibitem{haghighat2014fast}
M.~Haghighat and M.~A. Razian.
\newblock {Fast-FMI: non-reference image fusion metric}.
\newblock In {\em Application of Information and Communication Technologies
  (AICT), 2014 IEEE 8th International Conference on}, pages 1--3. IEEE, 2014.

\bibitem{he2016deep}
K.~He, X.~Zhang, S.~Ren, and J.~Sun.
\newblock {Deep residual learning for image recognition}.
\newblock In {\em Proceedings of the IEEE conference on computer vision and
  pattern recognition}, pages 770--778, 2016.

\bibitem{hossny2008comments}
M.~Hossny, S.~Nahavandi, and D.~Creighton.
\newblock {Comments on'Information measure for performance of image fusion'}.
\newblock {\em Electronics letters}, 44(18):1066--1067, 2008.

\bibitem{jiang2019hierarchical}
M.-x. Jiang, C.~Deng, J.-s. Shan, Y.-y. Wang, Y.-j. Jia, and X.~Sun.
\newblock {Hierarchical multi-modal fusion FCN with attention model for RGB-D
  tracking}.
\newblock {\em Information Fusion}, 50:1--8, 2019.

\bibitem{vot2019rgbt}
M.~Kristan, J.~Matas, A.~Leonardis, M.~Felsberg, R.~Pflugfelder, J.-K.
  Kamarainen, L.~Cehovin~Zajc, O.~Drbohlav, A.~Lukezic, A.~Berg, et~al.
\newblock {The seventh visual object tracking vot2019 challenge results}.
\newblock In {\em Proceedings of the IEEE International Conference on Computer
  Vision Workshops}, pages 0--0, 2019.

\bibitem{kumar2013multifocus}
B.~S. Kumar.
\newblock {Multifocus and multispectral image fusion based on pixel
  significance using discrete cosine harmonic wavelet transform}.
\newblock {\em Signal, Image and Video Processing}, 7(6):1125--1143, 2013.

\bibitem{kumar2015image}
B.~S. Kumar.
\newblock {Image fusion based on pixel significance using cross bilateral
  filter}.
\newblock {\em Signal, image and video processing}, 9(5):1193--1204, 2015.

\bibitem{lan2018modality}
X.~Lan, M.~Ye, S.~Zhang, H.~Zhou, and P.~C. Yuen.
\newblock {Modality-correlation-aware sparse representation for RGB-infrared
  object tracking}.
\newblock {\em Pattern Recognition Letters}, 2018.

\bibitem{li2019rgb}
C.~Li, X.~Liang, Y.~Lu, N.~Zhao, and J.~Tang.
\newblock {RGB-T object tracking: benchmark and baseline}.
\newblock {\em Pattern Recognition}, 96:106977, 2019.

\bibitem{li2019codes}
H.~Li.
\newblock Codes for {MDLatLRR}.
\newblock 2019.
\newblock \url{https://github.com/hli1221/imagefusion_mdlatlrr}.

\bibitem{li2017multi}
H.~Li and X.-J. Wu.
\newblock {Multi-focus image fusion using dictionary learning and low-rank
  representation}.
\newblock In {\em International Conference on Image and Graphics}, pages
  675--686. Springer, 2017.

\bibitem{li2018densefuse}
H.~Li and X.-J. Wu.
\newblock {DenseFuse: A Fusion Approach to Infrared and Visible Images}.
\newblock {\em IEEE Transactions on Image Processing}, 28(5):2614--2623, 2019.

\bibitem{li2019infrared}
H.~Li, X.-J. Wu, and T.~S. Durrani.
\newblock {Infrared and Visible Image Fusion with ResNet and zero-phase
  component analysis}.
\newblock {\em Infrared Physics \& Technology}, page 103039, 2019.

\bibitem{li2018infrared}
H.~Li, X.-J. Wu, and J.~Kittler.
\newblock {Infrared and Visible Image Fusion using a Deep Learning Framework}.
\newblock In {\em 2018 24th International Conference on Pattern Recognition
  (ICPR)}, pages 2705--2710. IEEE, 2018.

\bibitem{li2017pixel}
S.~Li, X.~Kang, L.~Fang, J.~Hu, and H.~Yin.
\newblock {Pixel-level image fusion: A survey of the state of the art}.
\newblock {\em Information Fusion}, 33:100--112, 2017.

\bibitem{liu2017infrared}
C.~Liu, Y.~Qi, and W.~Ding.
\newblock {Infrared and visible image fusion method based on saliency detection
  in sparse domain}.
\newblock {\em Infrared Physics \& Technology}, 83:94--102, 2017.

\bibitem{liu2010robust}
G.~Liu, Z.~Lin, and Y.~Yu.
\newblock {Robust subspace segmentation by low-rank representation}.
\newblock In {\em Proceedings of the 27th international conference on machine
  learning (ICML-10)}, pages 663--670, 2010.

\bibitem{liu2011latent}
G.~Liu and S.~Yan.
\newblock {Latent low-rank representation for subspace segmentation and feature
  extraction}.
\newblock In {\em Computer Vision (ICCV), 2011 IEEE International Conference
  on}, pages 1615--1622. IEEE, 2011.

\bibitem{liu2017multi}
Y.~Liu, X.~Chen, H.~Peng, and Z.~Wang.
\newblock {Multi-focus image fusion with a deep convolutional neural network}.
\newblock {\em Information Fusion}, 36:191--207, 2017.

\bibitem{liu2016image}
Y.~Liu, X.~Chen, R.~K. Ward, and Z.~J. Wang.
\newblock {Image fusion with convolutional sparse representation}.
\newblock {\em IEEE signal processing letters}, 23(12):1882--1886, 2016.

\bibitem{lu2014infrared}
X.~Lu, B.~Zhang, Y.~Zhao, H.~Liu, and H.~Pei.
\newblock {The infrared and visible image fusion algorithm based on target
  separation and sparse representation}.
\newblock {\em Infrared Physics \& Technology}, 67:397--407, 2014.

\bibitem{luo2017image}
X.~Luo, Z.~Zhang, B.~Zhang, and X.~Wu.
\newblock {Image fusion with contextual statistical similarity and
  nonsubsampled shearlet transform}.
\newblock {\em IEEE Sensors Journal}, 17(6):1760--1771, 2017.

\bibitem{ma2016infrared}
J.~Ma, C.~Chen, C.~Li, and J.~Huang.
\newblock {Infrared and visible image fusion via gradient transfer and total
  variation minimization}.
\newblock {\em Information Fusion}, 31:100--109, 2016.

\bibitem{ma2020infrared}
J.~Ma, P.~Liang, W.~Yu, C.~Chen, X.~Guo, J.~Wu, and J.~Jiang.
\newblock {Infrared and visible image fusion via detail preserving adversarial
  learning}.
\newblock {\em Information Fusion}, 54:85--98, 2020.

\bibitem{ma2019fusiongan}
J.~Ma, W.~Yu, P.~Liang, C.~Li, and J.~Jiang.
\newblock {FusionGAN: A generative adversarial network for infrared and visible
  image fusion}.
\newblock {\em Information Fusion}, 48:11--26, 2019.

\bibitem{ma2017infrared}
J.~Ma, Z.~Zhou, B.~Wang, and H.~Zong.
\newblock {Infrared and visible image fusion based on visual saliency map and
  weighted least square optimization}.
\newblock {\em Infrared Physics \& Technology}, 82:8--17, 2017.

\bibitem{ma2015perceptual}
K.~Ma, K.~Zeng, and Z.~Wang.
\newblock {Perceptual quality assessment for multi-exposure image fusion}.
\newblock {\em IEEE Transactions on Image Processing}, 24(11):3345--3356, 2015.

\bibitem{pang2012multifocus}
H.~Pang, M.~Zhu, and L.~Guo.
\newblock {Multifocus color image fusion using quaternion wavelet transform}.
\newblock In {\em Image and Signal Processing (CISP), 2012 5th International
  Congress on}, pages 543--546. IEEE, 2012.

\bibitem{prabhakar2017deepfuse}
K.~R. Prabhakar, V.~S. Srikar, and R.~V. Babu.
\newblock {Deepfuse: A deep unsupervised approach for exposure fusion with
  extreme exposure image pairs}.
\newblock In {\em 2017 IEEE International Conference on Computer Vision (ICCV).
  IEEE}, pages 4724--4732, 2017.

\bibitem{rao1997fibre}
Y.-J. Rao.
\newblock {In-fibre Bragg grating sensors}.
\newblock {\em Measurement science and technology}, 8(4):355, 1997.

\bibitem{roberts2008assessment}
J.~W. Roberts, J.~A. Van~Aardt, and F.~B. Ahmed.
\newblock {Assessment of image fusion procedures using entropy, image quality,
  and multispectral classification}.
\newblock {\em Journal of Applied Remote Sensing}, 2(1):023522, 2008.

\bibitem{sattar1997image}
F.~Sattar, L.~Floreby, G.~Salomonsson, and B.~Lovstrom.
\newblock {Image enhancement based on a nonlinear multiscale method}.
\newblock {\em IEEE transactions on image processing}, 6(6):888--895, 1997.

\bibitem{sheikh2006image}
H.~R. Sheikh and A.~C. Bovik.
\newblock {Image information and visual quality}.
\newblock {\em IEEE Transactions on image processing}, 15(2):430--444, 2006.

\bibitem{shrinidhi2018ir}
V.~Shrinidhi, P.~Yadav, and N.~Venkateswaran.
\newblock {IR and Visible Video Fusion for Surveillance}.
\newblock In {\em 2018 International Conference on Wireless Communications,
  Signal Processing and Networking (WiSPNET)}, pages 1--6. IEEE, 2018.

\bibitem{simonyan2014very}
K.~Simonyan and A.~Zisserman.
\newblock {Very deep convolutional networks for large-scale image recognition}.
\newblock {\em arXiv preprint arXiv:1409.1556}, 2014.

\bibitem{tno2018}
A.~Toet.
\newblock {TNO Image Fusion Dataset}, 2014.
\newblock \url{https://figshare.com/articles/TN_Image_Fusion_Dataset/1008029}.

\bibitem{vishwakarma2018image}
A.~Vishwakarma and M.~Bhuyan.
\newblock {Image Fusion Using Adjustable Non-subsampled Shearlet Transform}.
\newblock {\em IEEE Transactions on Instrumentation and Measurement}, 2018.

\bibitem{wang2014eggdd}
L.~Wang, B.~Li, and L.-F. Tian.
\newblock {EGGDD: An explicit dependency model for multi-modal medical image
  fusion in shift-invariant shearlet transform domain}.
\newblock {\em Information Fusion}, 19:29--37, 2014.

\bibitem{wright2008robust}
J.~Wright, A.~Y. Yang, A.~Ganesh, S.~S. Sastry, and Y.~Ma.
\newblock {Robust face recognition via sparse representation}.
\newblock {\em IEEE transactions on pattern analysis and machine intelligence},
  31(2):210--227, 2008.

\bibitem{wu2015object}
Y.~Wu, J.~Lim, and M.-H. Yang.
\newblock {Object tracking benchmark}.
\newblock {\em IEEE Transactions on Pattern Analysis and Machine Intelligence},
  37(9):1834--1848, 2015.

\bibitem{trwotrackers}
T.~Xu.
\newblock {The codes of LADCF and GFSDCF}, 2019.
\newblock \url{https://github.com/XU-TIANYANG}.

\bibitem{xu2019an}
T.~Xu, Z.-H. Feng, X.-J. Wu, and J.~Kittler.
\newblock {An Accelerated Correlation Filter Tracker}.
\newblock {\em Pattern Recognition accepted. arXiv preprint arXiv:1912.02854},
  2019.

\bibitem{xu2019joint}
T.~Xu, Z.-H. Feng, X.-J. Wu, and J.~Kittler.
\newblock {Joint group feature selection and discriminative filter learning for
  robust visual object tracking}.
\newblock In {\em Proceedings of the IEEE International Conference on Computer
  Vision}, pages 7950--7960, 2019.

\bibitem{xu2019learning}
T.~Xu, Z.-H. Feng, X.-J. Wu, and J.~Kittler.
\newblock {Learning Adaptive Discriminative Correlation Filters via Temporal
  Consistency preserving Spatial Feature Selection for Robust Visual Object
  Tracking}.
\newblock {\em IEEE Transactions on Image Processing}, 2019.

\bibitem{xydeas2000objective}
C.~Xydeas, , and V.~Petrovic.
\newblock {Objective image fusion performance measure}.
\newblock {\em Electronics letters}, 36(4):308--309, 2000.

\bibitem{yang2010image}
S.~Yang, M.~Wang, L.~Jiao, R.~Wu, and Z.~Wang.
\newblock {Image fusion based on a new contourlet packet}.
\newblock {\em Information Fusion}, 11(2):78--84, 2010.

\bibitem{yin2017novel}
M.~Yin, P.~Duan, W.~Liu, and X.~Liang.
\newblock {A novel infrared and visible image fusion algorithm based on
  shift-invariant dual-tree complex shearlet transform and sparse
  representation}.
\newblock {\em Neurocomputing}, 226:182--191, 2017.

\bibitem{zhang2013dictionary}
Q.~Zhang, Y.~Fu, H.~Li, and J.~Zou.
\newblock {Dictionary learning method for joint sparse representation-based
  image fusion}.
\newblock {\em Optical Engineering}, 52(5):057006, 2013.

\bibitem{zhang2020ifcnn}
Y.~Zhang, Y.~Liu, P.~Sun, H.~Yan, X.~Zhao, and L.~Zhang.
\newblock {IFCNN: A general image fusion framework based on convolutional
  neural network}.
\newblock {\em Information Fusion}, 54:99--118, 2020.

\bibitem{zong2017medical}
J.-j. Zong and T.-s. Qiu.
\newblock {Medical image fusion based on sparse representation of classified
  image patches}.
\newblock {\em Biomedical Signal Processing and Control}, 34:195--205, 2017.

\end{thebibliography}
}

\begin{IEEEbiography}[{\includegraphics[width=1in,height=1.25in,clip,keepaspectratio]{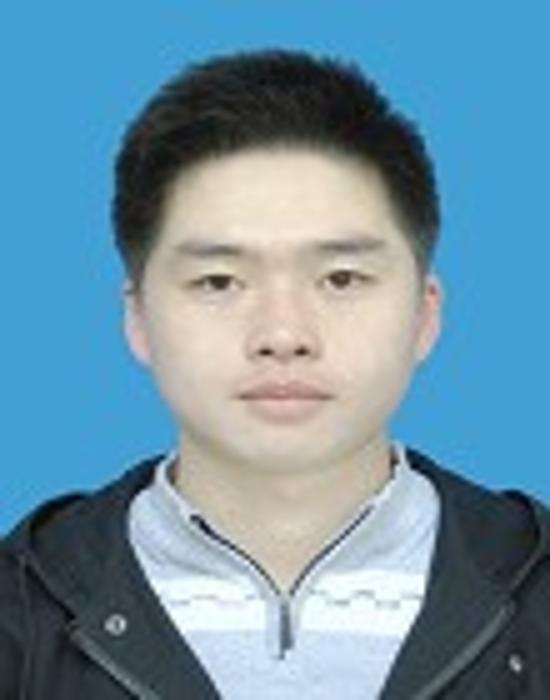}}]{Hui Li}
received the B.Sc. degree in School of Internet of Things Engineering from Jiangnan University, China, in 2015. He is currently a PhD student at the Jiangsu Provincial Engineerinig Laboratory of Pattern Recognition and Computational Intelligence, Jiangnan University. His research interests include image fusion, machine learning and deep learning.
\end{IEEEbiography}

\begin{IEEEbiography}[{\includegraphics[width=1in,height=1.25in,clip,keepaspectratio]{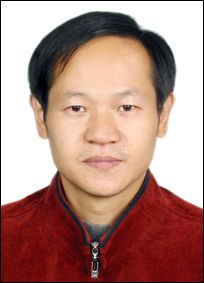}}]{Xiao-Jun Wu}
received the B.Sc. degree in mathematics from Nanjing Normal University, Nanjing, China,
in 1991, and the M.S. degree and Ph.D. degree
in pattern recognition and intelligent system from
the Nanjing University of Science and Technology, Nanjing, in 1996 and 2002, respectively. From
1996 to 2006, he taught at the School of Electronics
and Information, Jiangsu University of Science and
Technology, where he was promoted to Professor.

He has been with the School of Information Engineering, Jiangnan University since 2006, where he
is a Professor of pattern recognition and computational intelligence. He was a Visiting
Researcher with the Centre for Vision, Speech, and Signal Processing (CVSSP), University of Surrey, U.K. from 2003 to 2004. He has published over 300 papers in his fields of research.
His current research interests include pattern recognition, computer vision, and
computational intelligence. He was a Fellow of the International Institute for
Software Technology, United Nations University, from 1999 to 2000. He was
a recipient of the Most Outstanding Postgraduate Award from the Nanjing
University of Science and Technology.
\end{IEEEbiography}

\begin{IEEEbiography}[{\includegraphics[width=1in,height=1.25in,clip,keepaspectratio]{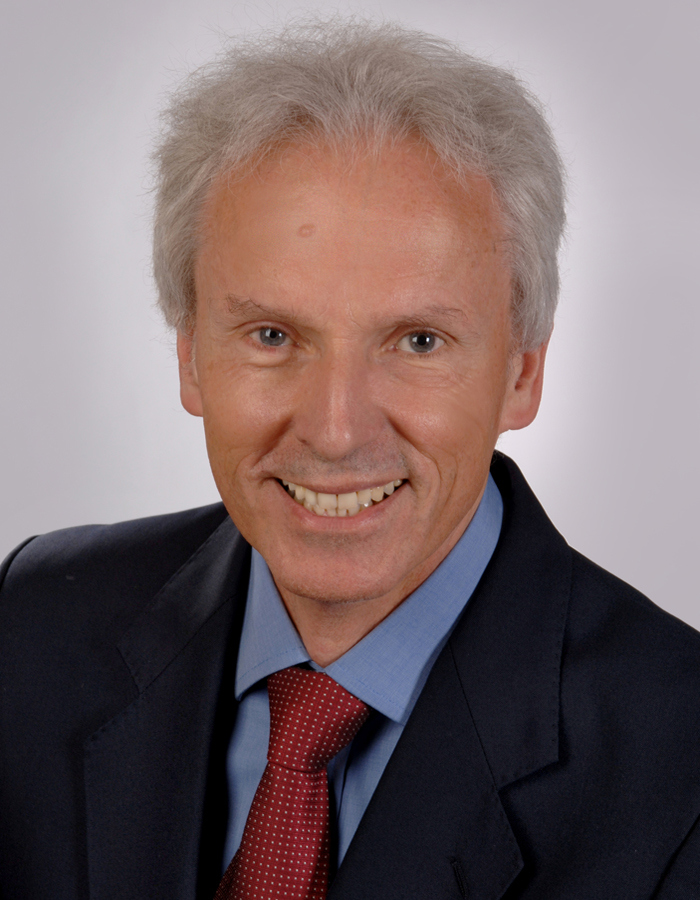}}]{Josef Kittler}
(M’74-LM’12) received the B.A.,
Ph.D., and D.Sc. degrees from the University of
Cambridge, in 1971, 1974, and 1991, respectively.
He is a distinguished Professor of Machine Intelligence
at the Centre for Vision, Speech and Signal
Processing, University of Surrey, Guildford, U.K.
He conducts research in biometrics, video and image
database retrieval, medical image analysis, and
cognitive vision. He published the textbook Pattern
Recognition: A Statistical Approach and over 700
scientific papers. His publications have been cited
more than 60,000 times (Google Scholar).

He is series editor of Springer Lecture Notes on Computer Science. He
currently serves on the Editorial Boards of Pattern Recognition Letters, Pattern
Recognition and Artificial Intelligence, Pattern Analysis and Applications. He
also served as a member of the Editorial Board of IEEE Transactions on
Pattern Analysis and Machine Intelligence during 1982-1985. He served on
the Governing Board of the International Association for Pattern Recognition
(IAPR) as one of the two British representatives during the period 1982-2005,
President of the IAPR during 1994-1996.
\end{IEEEbiography}

\end{document}